\documentclass[preprint, review]{elsarticle}
\usepackage[margin=1in]{geometry}
\usepackage{epsfig}
\usepackage{amssymb}
\usepackage{amsmath}
\usepackage{multirow}
\usepackage{lipsum}
\usepackage{caption}
\usepackage{multicol}
\usepackage{longtable,ltxtable,booktabs}
\usepackage{lscape}
\usepackage{lipsum} % for dummy text
\usepackage{enumitem}
\usepackage{array}
\usepackage{color}
\usepackage[strings]{underscore}
\usepackage[caption = false]{subfig}
\usepackage{graphicx}

\usepackage{url}
\makeatletter
\g@addto@macro{\UrlBreaks}{\UrlOrds}
\makeatother

\newcolumntype{P}[1]{>{\endgraf\vspace*{-\baselineskip}}p{#1}}
\usepackage{forloop}


\journal{Pattern Recognition}

\begin{document}

\begin{frontmatter}



\title{RGB-D-based Action Recognition Datasets: A Survey}



\author[label1]{Jing Zhang\corref{mycorrespondingauthor}}
\cortext[mycorrespondingauthor]{Corresponding author.}
\ead{jz960@uowmail.edu.au}
\author[label1]{Wanqing Li}
\ead{wanqing@uow.edu.au}
\author[label1]{Philip O. Ogunbona}
\ead{philipo@uow.edu.au}
\author[label1]{Pichao Wang}
\ead{pw212@uowmail.edu.au}
\author[label1,label2]{Chang Tang}
\ead{tangchang@tju.edu.cn}

\address[label1]{School of Computing and Information Technology, University of Wollongong, NSW 2522, Australia}
\address[label2]{School of Electronic Information Engineering, Tianjin University, Tianjin 300072, China}

\begin{abstract}
%Human activity understanding from RGB-D data has attracted increased attention 
%since the first work reported in 2010. Over this period, many benchmark datasets 
%have been created to facilitate the development and evaluation of new 
%algorithms. This raises the question of which dataset to select and how to use 
%it in providing a fair and objective comparative evaluation against 
%state-of-the-art methods. To address this issue, this paper provides a 
%comprehensive review of the most commonly used action recognition related RGB-D 
%video datasets, including 26 single-view datasets, 10 multi-view datasets, and 7 
%multi-person datasets. Our contribution of detailed information and analysis 
%of these datasets is a useful resource in guiding insightful selection of 
%datasets for future research. In addition, the issues with current algorithm 
%evaluation vis-\'{a}-vis limitations of the available datasets and evaluation protocol are 
%also highlighted; resulting in a number of recommendations for future collection 
%of datasets and future use of evaluation protocols.
Human action recognition from RGB-D (Red, Green, Blue and Depth) data has attracted increasing attention 
since the first work reported in 2010. Over this period, many benchmark datasets 
have been created to facilitate the development and evaluation of new 
algorithms. This raises the question of which dataset to select and how to use 
it in providing a fair and objective comparative evaluation against 
state-of-the-art methods. To address this issue, this paper provides a 
comprehensive review of the most commonly used action recognition related RGB-D 
video datasets, including 27 single-view datasets, 10 multi-view datasets, and 7 
multi-person datasets. The detailed information and analysis 
of these datasets is a useful resource in guiding insightful selection of 
datasets for future research. In addition, the issues with current algorithm 
evaluation 
vis-\'{a}-vis limitations of the available datasets and evaluation protocols are 
also highlighted; resulting in a number of recommendations for collection 
of new datasets and use of evaluation protocols.

 \end{abstract} 
% REMOVED %In addition, the issues with current algorithm evaluation 
% vis-\'{a}-vis % limitations of the available datasets and evaluation protocol 
% are also % highlighted. 
% %To overcome these limitations, a large scale dataset is
% proposed along with a set of evaluation protocols. Several experiments using 
% two % state-of-the-art algorithms are conducted on the combined dataset and 
% results % show that current algorithms only work well within specific dataset 
% and % evaluation protocol. Furthermore, the experiments also verify the 
% effectiveness % of the large dataset on evaluation of algorithms from different 
% perspectives. 


\end{frontmatter}

% \tableofcontents %help to visualize paper organization
\section{Introduction}
\label{sec:introduction}
%Current computer vision systems poorly compare with the performance 
%of the human visual system in many tasks. For example, accurate recognition 
%of human actions is still a challenging task for computer vision systems largely 
%because they depend of 2D RGB images. The introduction of low-cost integrated  
%depth sensors (such as Microsoft Kinect~\texttrademark) that can capture both 
%RGB and depth information has provided additional opportunity for researchers to 
%improve the recognition rate. This development has led to the generation of a 
%number of action datasets captured by depth sensor for evaluating the 
%performance of new action recognition algorithms.

Human action recognition is an active research topic in Computer Vision. Prior to the release of Microsoft Kinect~\texttrademark, research has mainly focused on learning and recognizing actions from conventional two-dimensional (2D) video~\cite{Vishwakarma2013, Lim2015, Wang2003, Guo2014}. There are many publicly available 2D video datasets dedicated to action recognition. Review papers categorizing and summarizing their characteristics are available to help researchers in evaluating their algorithms~\cite{Hassner2013, Chaquet2013, Ruffieux2014}.
%~\cite{Poppe2010, Aggarwal2011, Vishwakarma2013, Weinland2011, Hassner2013, Chaquet2013, Ruffieux2014, Lim2015, Wang2003, Guo2014}. 
The introduction of low-cost integrated depth sensors (such as Microsoft Kinect~\texttrademark) that can capture both RGB (red, green and blue) video and depth (D) information has significantly advanced the research of human action recognition.  Since the first work reported in 2010~\cite{Li2010}, many benchmark datasets have been created to facilitate the development and evaluation of new action recognition algorithms. However, available RGB-D-based datasets have insofar only been briefly summarized or enumerated without comprehensive coverage and in-depth analysis in the survey papers, such as~\cite{Aggarwal2014,Lun2015},
%~\cite{Ye2013,Aggarwal2014,Chen2013a,Ruffieux2014,Lun2015}, 
that mainly focus on the development of RGB-D-based action recognition algorithms. The lack of comprehensive reviews on RGB-D datasets motivated the focus of this paper.

Datasets are important for the rapid development and objective evaluation and comparison of algorithms. To this end, they should be carefully created or selected to ensure effective evaluation of the validity and efficacy of any algorithm under investigation. The evaluation of each task-specific algorithm depends not only on the underlying methods but also on the factors captured by each dataset. However, it is currently difficult to select the most appropriate dataset from among the many Kinect sensor captured RGB-D datasets available and establish the most appropriate evaluation protocol. There is also the possibility of creating a new but redundant dataset because of the lack of comprehensive survey on what is available. This paper fills this gap by providing comprehensive summaries and analysis of existing RGB-D action datasets and the evaluation protocols that have been used in association with these datasets. 

The paper focuses on action and activity datasets.  ``Gesture datasets" are excluded from this survey since, unlike actions and activities that usually involve motion of the entire human body,  gesture involves only hand movement and gesture recognition is often considered as a research topic independent of action and activity recognition. For details of the available gesture datasets, readers are referred to the survey paper by Ruffieux et al.~\cite{Ruffieux2014}.

This rest of the survey is organized as follows.
%Section~\ref{sec:related} reviews previous 
%work on survey of action recognition related datasets from the viewpoint of 
%2D video datasets and RGB-D video datasets. 
Section~\ref{sec:datasets} summarises characteristics of publicly available and 
commonly used RGB-D datasets; the summaries (44 in total) are categorised under 
\textit{single-view activity/action datasets}, \textit{multi-view 
action/activity datasets} and \textit{interaction/multi-person activity 
datasets}. Section~\ref{sec:analysis} provides a comparative analysis of the 
reviewed datasets with regard to the applications, complexity, state-of-the-art 
results, and commonly employed evaluation protocols. In addition, some 
recommendations are provided to aid the future usage of datasets and evaluation 
protocols. Discussions on the limitations of current RGB-D action datasets and 
commonly used evaluation methods are presented in Section~\ref{sec:Discussion}. 
At the same time, we provide some recommendations on requirements for
future creation of datasets and selection of evaluation protocols.
In Section~\ref{sec:conclusion}, a brief conclusion is drawn.

\section{RGB-D Action/Activity Datasets}
\label{sec:datasets}
%%Currently, there is no literature that provides a comprehensive review of RGB-D 
%% datasets for action recognition tasks. 
%In this section we summarize most of the publicly available RGB-D action datasets 
%captured by depth sensors along with comprehensive description and 
%specifications. The detailed description of each dataset includes the creation 
%date, creation institution, number of actions, number of subjects involved, 
%action repetition times, action classes, total number of video samples, capture settings, background and environment. 
%%Where it is informative we show some sample frames or capture environment settings.

This section summarizes most of the publicly available RGB-D action datasets, including the creation 
date, creation institution, number of actions, number of subjects involved, 
action repetition times, action classes, total number of video samples, capture settings, background and environment. 

The datasets are categorized into three classes namely: 
\textit{single-view action/activity}, \textit{multi-view action/activity}, and 
\textit{human-human interaction/multi-person activity}. In the single-view 
action/activity datasets, each action is captured from a single specific view 
point, while in the multi-view action/activity datasets, two or more 
view points of each action are captured. Note that in both single-view and 
multi-view datasets, each action/activity is performed by one actor at a time. 
The human-human interaction/multi-person activity datasets consist of 
interactions between two people or activities performed by multiple persons.

%First, we clarify some terminologies used in this section. A ``gesture 
%dataset'' contains only hand movement. An ``activity dataset'' contains whole 
%body movement performed by human actor for a certain period of time. If the 
%duration of the action is relatively short it is referred to as ``action 
%dataset''.

%This paper focuses on datasets used in the development and evaluation of human 
%action recognition algorithms and as such we only review action and activity 
%datasets.  We consider three categories of RGB-D datasets namely: 
%\textit{single-view action/activity}, \textit{multi-view action/activity}, and 
%\textit{human-human interaction/multi-person activity}. In the single-view 
%action/activity datasets, each action is captured from a single specific view 
%point, while in the multi-view action/activity datasets, two or more 
%view points of each action are captured. Note that in both single-view and 
%multi-view datasets, each action/activity is performed by one actor at a time. 
%The human-human interaction/multi-person activity datasets consist of 
%interactions between two people or activities performed by multiple persons. Notice that ``gesture datasets" are excluded from this survey since, unlike actions and activities that usually involve motion of entire human body,  gesture involves only hand movement and gesture recognition is usually considered as an research topic independent from action and activity recognition.

\subsection{Single-view action/activity datasets}
%In single-view action/activity dataset, each action or activity is performed by 
% a single person from a single viewpoint. 
%This section reviews most commonly used and publicly available single-view 
%datasets. The description of these datasets is provided in chronological 
%order. 
Table~\ref{table:Basicsingle} is a list  
summarizing the basic specifications of single view action/activity 
datasets in descending order of citation frequency. 
\subsubsection{MSR-Action3D}
MSR-Action3D~\cite{Li2010}(\url{http://research.microsoft.com/en-us/um/people/zliu/ActionRecoRsrc/}) is the first public benchmark RGB-D action 
dataset collected by Microsoft Research Redmond and University of Wollongong in 2010. The 
dataset contains 20 actions: \textit{high arm wave}, \textit{horizontal arm 
wave}, \textit{hammer}, \textit{hand catch}, \textit{forward punch}, 
\textit{high throw}, \textit{draw x}, \textit{draw tick}, \textit{draw circle}, 
\textit{hand clap}, \textit{two hand wave}, \textit{side-boxing}, \textit{bend}, 
\textit{forward kick}, \textit{side kick}, \textit{jogging}, \textit{tennis 
serve}, \textit{golf swing}, \textit{pickup and throw}. 
%In Figure~\ref{fig:MSR-Action3D}, some sample depth frames are shown, where the depth value is encoded with Red, Green, Blue color information. 
Ten subjects performed these actions three times. All the 
videos were recorded from a fixed point of view and the subjects were facing the 
camera while performing the actions. The background of the dataset 
was removed by some post-processing. Specifically, if an action needs to be 
performed with one arm or one leg, the actors were required to perform it using 
right arm or leg. The data are provided as segmented samples and the sample 
file names provide the information of action types, subject ID and number of 
repetitions. 

%\begin{figure}[htbp]
%\centering
%\subfloat[Draw tick]{
%\begin{minipage}[t]{0.45\textwidth}
%\centering
%\includegraphics[keepaspectratio=true, scale=0.3]{./images/MSRAction3D1.jpg}
%\end{minipage}
%}
%\subfloat[Tennis serve]{
%\begin{minipage}[t]{0.49\textwidth}
%\centering
%\includegraphics[keepaspectratio=true, scale=0.33]{./images/MSRAction3D2.jpg}
%\end{minipage}
%}
%\caption{Sample frames of the MSR-Action3D dataset~\cite{Li2010}.}
%\label{fig:MSR-Action3D}
%\end{figure}

\subsubsection{RGBD-HuDaAct}
RGBD-HuDaAct~\cite{Ni2011}(\url{http://adsc.illinois.edu/sites/default/files/files/ADSC-RGBD-dataset-download-instructions.pdf}) was collected by Advanced Digital Sciences Center 
Singapore in 2011. Compared to MSR-Action3D dataset, this dataset consists of 
fewer actions (12 actions) and performed by more subjects (30 subjects). The 
action types are also different from MSR-Action3D dataset.  This dataset
focuses on human daily activities, such as \textit{make a phone call}, 
\textit{mop the floor}, \textit{enter the room}, \textit{exit the room}, 
\textit{go to bed}, 
\textit{get up, eat meal}, \textit{drink water}, \textit{sit down}, 
\textit{stand up}, t\textit{ake off the jacket}, and \textit{put on the jacket}. 
Each actor performed 2-4 repetitions of each action. The background is also 
fixed as the camera was fixed when recording. However, there was no 
restriction on which leg or hand was used in the actions and the dataset 
contains human-object interaction. 
%Figure~\ref{fig:RGBD-HuDaAct} gives the sample frames of RGB and depth video of the $12$ actions.
%\begin{figure}[ht!]
%   \centering
%   \includegraphics[keepaspectratio=true, scale =0.4]{./images/RGBD-HuDaAct.jpg} 
%  
%   \caption{Sample frames of the RGBD-HuDaAct dataset~\cite{Ni2011}.}
%   \label{fig:RGBD-HuDaAct}
%\end{figure}
\subsubsection{CAD-60}
CAD-60 dataset~\cite{Sung2011}(\url{http://pr.cs.cornell.edu/humanactivities/data.php}) was captured by Cornell University in 
2011, motivated by the fact that true daily activities rarely occur in 
structured environments. Hence, the actions were performed within
uncontrolled background. Twelve distinctive activities were performed within 5 
environments: 
bathroom (\textit{rinsing mouth}, \textit{brushing teeth}, \textit{wearing 
contact lens}), bedroom 
(\textit{talking on the phone}, \textit{drinking water}, \textit{opening pill 
container}), kitchen 
(cooking (\textit{chopping}), cooking (\textit{stirring}), \textit{drinking 
water}, \textit{opening pill 
container}), living room (\textit{talking on the phone}, \textit{drinking 
water}, \textit{talking on couch}, 
\textit{relaxing on couch}), office (\textit{talking on the phone}, 
\textit{writing on 
whiteboard}, 
\textit{drinking water}, \textit{working on computer}). Four subjects performed 
all the activities and one of the subjects is left-handed. To determine 
whether test algorithms can distinguish the desired activities from other 
randomly performed activities, additional random activity was collected, which 
contains a series of random 
movements that is different from any of other 12 activities in the dataset. In 
the original paper, this random activity was only used at testing stage. 
% Some sample color images can be found in Figure~\ref{fig:CAD-60}.
% \begin{figure}[htbp!]
%    \centering
%     \includegraphics[keepaspectratio=true, scale =0.35]{./images/CAD60.png}
%    \caption{Sample frames of the CAD-60 dataset~\cite{Sung2011}.}
%    \label{fig:CAD-60}
% \end{figure}
\subsubsection{MSRC-12}
MSRC-12 dataset~\cite{Fothergill2012}(\url{http://research.microsoft.com/en-us/um/cambridge/projects/msrc12/}) was collected by Microsoft 
Research Cambridge and University of Cambridge in 2012. Although it is 
sometimes referred as gesture dataset, the movements involved whole body, so we 
categorize it as action/activity dataset. Two main goals motivated the 
collection of this dataset: first, to test whether semiotic modality of 
instructions for collecting data will affect the performance of the 
recognition system and, second, to determine whether the type of gesture 
makes a difference in the effect of modality. So, there are two types of 
gestures: Iconic gestures (\textit{Crouch} or \textit{hide}, \textit{Shoot a 
pistol}, \textit{Throw an object}, \textit{Change weapon}, \textit{Kick}, 
and \textit{Put on night vision goggles}) and Metaphoric gestures 
(\textit{Start Music/Raise Volume (of music)}, \textit{Navigate to next menu}, 
\textit{Wind up the music}, \textit{Take a bow to end music session}, 
\textit{Protest the music}, and \textit{Move up the tempo of the song}). The 
authors provided three familiar and easy to prepare instruction modalities and 
their combinations to the participants. The modalities are (1) descriptive text 
breaking down the performance kinematics, (2) an ordered series of static 
images 
of a person performing the gesture with arrows annotating as appropriate, and 
(3) video (dynamic images) of a person performing the gesture. 
%Some examples of the descriptive text and static image instruction for iconic gesture are shown in Table~\ref{tab:MSRC-12}. 
There are 30 participants in total and for each 
gesture, the data were collected as: Text 
(10 people), Images (10 people), Video (10 people), Video with text (10 people), 
Images with text (10 people). The dataset was captured using 
Kinect~\texttrademark sensor and only the skeleton data are made available. 

\subsubsection{MSRDailyActivity3D}
MSRDailyActivity3D Dataset~\cite{Wang2012}(\url{http://research.microsoft.com/en-us/um/people/zliu/ActionRecoRsrc/}) was collected by  Microsoft and the 
Northwestern University in 2012 and focused on daily activities. The motivation 
was to cover human daily activities in the living 
room. There are 16 activity types: \textit{drink}, \textit{eat}, \textit{read 
book}, \textit{call cellphone}, \textit{write 
on a paper}, \textit{use laptop}, \textit{use vacuum cleaner}, \textit{cheer 
up}, \textit{sit still}, \textit{toss paper}, 
\textit{play game}, \textit{lay down on sofa}, \textit{walk, play guitar}, 
\textit{stand 
up}, \textit{sit down}. 
% Figure~\ref{fig:MSRDailyActivity} shows 
% sample depth frames of MSRDailyActivity3D dataset. 
The actions were performed by 10 actors while sitting on 
the sofa or standing close to the sofa. The camera was fixed in front of the 
sofa. In addition to depth data, skeleton data are also recorded, but the joint 
positions extracted by the tracker are very noisy due to the actors being either sitting on or standing close to the sofa. 
%% \begin{figure}[htbp!]
%%    \centering
%% %    \includegraphics[keepaspectratio=true, scale =0.35]{./images/MSRDailyActivity.jpg}
%%    \caption{Sample frames of MSR Daily Activity dataset. From~\cite{Wang2012}}
%%    \label{fig:MSRDailyActivity}
%% \end{figure}
\subsubsection{UTKinect}
UTKinect dataset~\cite{Xia2012}(\url{http://cvrc.ece.utexas.edu/KinectDatasets/HOJ3D.html}) was collected by the University of Texas at 
Austin in 2012. Ten types of human actions were performed twice by 10 subjects. 
The actions include \textit{walk}, \textit{sit down}, \textit{stand up}, 
\textit{pick up}, \textit{carry}, \textit{throw}, 
\textit{push}, \textit{pull}, \textit{wave}, \textit{clap hands}. 
% Sample frames can be found in figure~\ref{fig:UTKinect}. 
The subjects performed the actions from a variety of 
views. An added difficulty of recognition was afforded by the actions being 
performed with actor-dependent variability. Furthermore, human-object 
occlusions and body parts being out of the field of view added to the difficulty 
of the dataset in recognition tasks. Ground truth in terms of action labels and 
segmentation of sequences are provided. 
%% \begin{figure}[htbp!]
%%    \centering
%% %    \includegraphics[keepaspectratio=true, scale =0.5]{./images/UTKinect.png}
%%    \caption{Sample frames of UTKinect dataset. From~\cite{Xia2012}}
%%    \label{fig:UTKinect}
%% \end{figure}
\subsubsection{G3D}
Gaming 3D dataset (G3D)~\cite{Bloom2012,Bloom2013}(\url{http://dipersec.king.ac.uk/G3D/}) captured by 
Kingston University in 2012 focuses on real-time action recognition in gaming 
scenario. 
It contains 10 subjects performing 20 gaming actions: \textit{punch right}, 
\textit{punch left}, 
\textit{kick right}, \textit{kick left}, \textit{defend}, \textit{golf swing}, 
\textit{tennis serve}, \textit{throw bowling ball}, \textit{aim 
and fire gun}, \textit{walk}, \textit{run}, \textit{jump}, \textit{climb}, 
\textit{crouch}, \textit{steer a car}, \textit{wave}, \textit{flap}, and 
\textit{clap}. 
%Figure~\ref{fig:G3D} shows one frame sample of RGB, depth, and skeleton data. 
Each subject performed these actions thrice.
%In this dataset, depth data were stored in 16-bit greyscale images, 13 bits of 
% which were for depth data and the rest 3 bits stored the player's 
% identification. 
Two kinds of labels were provided as ground truth: the onset and 
offset of each action and, the peak frame of each action. In 
~\cite{Bloom2013}, the authors defined an \emph{action point} as a single time 
instance that an action is clear and all instances of that action can be 
uniquely identified. The peak frame provided in this dataset represents 
the action point indicated by the authors. This action point can be used for 
evaluating on-line action recognition algorithms.
%\begin{figure}[htbp!]
%   \centering
%   \includegraphics[keepaspectratio=true, scale =0.4]{./images/G3D.png}
%   \caption{One sample frame of the G3D dataset~\cite{Bloom2012}.}
%   \label{fig:G3D}
%\end{figure}
\subsubsection{DHA}
Depth-included Human Action video dataset (DHA)~\cite{Lin2012}(\url{http://mclab.citi.sinica.edu.tw/dataset/dha/dha.html}) was created by 
CITI in Academia Sinica. It contains 23 different actions: \textit{bend},
\textit{jack}, 
\textit{jump}, \textit{run}, \textit{side}, \textit{skip}, 
\textit{walk}, \textit{one-hand-wave}, 
\textit{two-hand-wave}, \textit{front-clap}, \textit{side-clap}, 
\textit{arm-swing}, 
\textit{arm-curl}, \textit{leg-kick}, \textit{leg-curl}, \textit{rod-swing},
\textit{golf-swing},
\textit{front-box}, \textit{side-box}, \textit{tai-chi}, \textit{pitch}, 
\textit{kick}. The first 10 
categories follow the same definitions as the Weizmann action 
dataset~\cite{Blank2005} and the 11th to 16th actions are extended categories. 
The 17th to 23rd are the categories of selected sport actions. 
%Figure~\ref{fig:DHA} illustrates sample frames of different actions involved in the dataset. 
The 23 actions were performed by 21 different individuals. All the actions were performed in 
one of three different scenes. Similarly to MSRAction3D dataset, the background 
information has been removed in the depth data.
%\begin{figure}[htbp!]
%   \centering
%    \includegraphics[keepaspectratio=true, scale =0.4]{./images/DHA.png}
%   \caption{Sample frames from the DHA dataset~\cite{Lin2012}.}
%   \label{fig:DHA}
%\end{figure}

\subsubsection{Falling Event Detection}
The Falling Event Detection dataset~\cite{Zhang2012a}(\url{http://media-lab.engr.ccny.cuny.edu/~zcy/}) was collected in 2012 by
City University of New York with the aim of creating a dataset 
for evaluating a newly proposed method for falling event detection and 
recognition. There are five activities related to falling event 
including \textit{standing, fall from standing, fall from sitting, sit on a 
chair}, and \textit{sit on floor}, captured using a RGB-D camera. The activities 
were performed by five different subjects under two different lighting 
environments (sufficient and insufficient illumination) resulting in 150 video 
sequences (100 videos under sufficient and 50 videos under insufficient 
illumination). The authors set aside a training set comprising 
50 videos which covers all 5 subjects and 5 types of activities performed under 
sufficient lighting. The remaining 100 video sequences (50 
for each condition) were set aside for testing. 
% The dataset is useful for 
% distinguishing falling event from other similar activities related to falling 
% such as sit on floor. It can also be used for user identification. 
% Figure~\ref{fig:FallingEvent} shows some sample frames 
% of Falling Event Detection dataset under different lighting environments
%% \begin{figure}[htbp!]
%%    \centering
%%     \includegraphics[keepaspectratio=true, scale=0.7]{./images/FallingEvent.png}
%%    \caption{Sample frames of Falling Event Detection dataset under different
%% lighting environments~\cite{Zhang2012a}.}
%%    \label{fig:FallingEvent}
%% \end{figure}
\subsubsection{MSRActionPair}
MSRActionPair dataset~\cite{Oreifej2013}(\url{http://www.cs.ucf.edu/~oreifej/HON4D.html}) was collected by University of Central 
Florida and Microsoft in 2013, and has two foci. First, 
the authors argue that many actions share similar motion cues; hence, relying 
only on motion information is insufficient for recognition. Second, considering 
motion and shape information independently is inefficient because they are 
correlated in an action sequence. As a result, they collected a dataset with 
pairs of actions; for example, \textit{pick up} and \textit{put down}. The 
action pairs share similar motion and shape cues but the relation between motion 
and shape is different. The background of the dataset was fixed, without 
occlusion and change of lighting. To perform well on this dataset, the 
algorithm needs to be able to capture the prominent cues of motion and shape 
jointly. In this dataset, ten subjects performed six pairs of actions 
twice: \textit{pick up a box/put down a box}, \textit{lift a box/place a box}, 
\textit{push a chair/pull a chair}, \textit{wear a hat/take off a hat}, 
\textit{put on a backpack/take off a backpack}, and \textit{stick a 
poster/remove a 
poster}. 
%% Figure~\ref{fig:MSR-ActionPair} shows some sample depth frames.
%% \begin{figure}[htbp!]
%%    \centering
%% %    \includegraphics[keepaspectratio=true, scale =0.6]{./images/MSR-ActionPair.png}
%%    \caption{Sample frames of MSR-ActionPair dataset. From~\cite{Oreifej2013}}
%%    \label{fig:MSR-ActionPair}
%% \end{figure}
\subsubsection{CAD-120}
CAD-120 dataset~\cite{Koppula2013a}(\url{http://pr.cs.cornell.edu/humanactivities/data.php}), collected by the Cornell University, 
focuses on high level activities and object interactions. This dataset contains 
10 high level activities performed by 4 subjects, and each activity was 
performed thrice with different objects. The high level activities 
include: \textit{making cereal}, \textit{taking medicine}, \textit{stacking 
objects}, \textit{unstacking objects}, 
\textit{microwaving food}, \textit{picking objects}, \textit{cleaning objects}, 
\textit{taking food}, \textit{arranging 
objects}, \textit{having a meal}. 
%The sample RGB and depth frames are shown in Figure~\ref{fig:CAD120}. 
The high level activities consist of a 
sequence of sub-activities. Different subjects performed the sub-activities 
over different length of time and, in different order and manner of execution. 
In addition, the 
subjects may perform the same activity with different objects. The 
backgrounds are also varied among actions. Based on above features, CAD-120 
dataset not only can be used for action recognition, but also can be 
used to evaluate some 
object detection and tracking algorithms. The dataset also provides some 
ground-truth, such as the bounding boxes of the objects involved in the 
 activities, sub-activity labels and object affordance labels. 
%% %10 sub-activity labels are: reaching, moving, pouring, eating, drinking, 
%% opening, placing, closing, scrubbing, null. 12 object affordance labels are: 
%% reachable, movable, pourable, pourto, containable, drinkable, openable, 
%% placeable, closable, scrubbable, scrubber, stationary.
%\begin{figure}[htbp!]
%   \centering
%   \includegraphics[keepaspectratio=true, scale =0.3]{./images/CAD120.png}
%   \caption{Sample frames of the CAD120 dataset~\cite{Koppula2013a}.}
%   \label{fig:CAD120}
%\end{figure}
\subsubsection{WorkoutSU-10 dataset}
WorkoutSU-10 dataset~\cite{Negin2013}(\url{http://vpa.sabanciuniv.edu/databases/WorkoutSU-10/}) was collected by Sabancı University in 
2013 and contains exercise actions selected by professional trainers for 
therapeutic purposes. There are 10 actions in total, namely \textit{SL Balance 
with Hip Flexion, SL Balance-Trunk Rotation (A2), Lateral Stepping, Thoracic 
Rotation – Bar on shoulder, Hip Adductor Stretch, Hip Adductor Stretch, DB 
Curl-to-Press, Freestanding Squats, Transverse Horizontal DB Punch, Transverse 
Horizontal DB Punch.} The performance instruction was the 
combination of an animated character performing the exercise and a subscripted 
text explaining the instructions. The RGB, depth, and skeleton data were all 
captured. Twelve subjects performed all the actions 10 times. 
There are 1200 action samples in total. The participants performed the action in 
front of a green screen, suggesting that the background of this dataset is 
clean.

%% \begin{figure}[htbp!]
%%    \centering
%% %    \includegraphics[keepaspectratio=true, scale =0.7]{./images/workoutsu.png}   
%%    \caption{Sample frames of WorkoutSU-10 dataset. (From http://vpa.sabanciuniv.edu/phpBB2/vpa\_views.php?s=31\&serial=36)}
%%    \label{fig:workoutsu}
%% \end{figure}

\subsubsection{Concurrent Action}
The concurrent action dataset~\cite{Wei2013a}(\url{http://www.stat.ucla.edu/~ping.wei/research/project/ConcurrentAction/ConcurrentAction.html}) was 
collected by Xi'an Jiaotong University and University of California, Los Angeles 
in 2013. This dataset focuses on action detection. Twelve actions were performed 
by several subjects in a sequential fashion. The actions are:\textit{ drink, 
make a call, turn on monitor, type on keyboard, fetch water, pour water, press 
button, pick up trash, throw trash, bend down, sit,} and \textit{stand}. 
Sixty-one long video sequences were captured. Each sequence contains 
several actions which are concurrent in the time and interact with others. 
The dataset is different from previously created dataset in that it 
contains multiple concurrent actions in each sequence and the actions 
semantically and temporally interact with each other. Only skeleton data 
format are available for this dataset. 
% Figure~\ref{fig:concurrent} is some sample 
% skeleton frames.
% 
%% \begin{figure}[htbp!]
%%    \centering
%% %    \includegraphics[keepaspectratio=true, scale =0.5]{./images/concurrent2.png}   
%%    \caption{Sample skeleton frames in concurrent dataset. From~\cite{Wei2013a}}
%%    \label{fig:concurrent}
%% \end{figure}
\subsubsection{IAS-lab Action}
IAS-lab Action dataset~\cite{Munaro2013,Munaro2013a}(\url{http://robotics.dei.unipd.it/actions/index.php/overview}) was collected by 
IAS Lab at the University of Padua in 2013. The authors claimed that in order 
to test as many different algorithms as possible, a dataset needs to contain 
sufficient variety of actions and number of people performing the 
actions. To this end, they captured 15 different actions performed by 12 
different people thrice. The actions are: \textit{check watch}, \textit{cross 
arms}, \textit{get up}, \textit{kick}, \textit{pick up}, 
\textit{point}, \textit{punch}, \textit{scratch head}, \textit{sit down}, 
\textit{standing}, \textit{throw from bottom up}, \textit{throw over 
head}, \textit{turn around}, \textit{walk}, and \textit{wave}. 
%Sample frames of each action are illustrated in Figure~\ref{fig:IAS}. 
The subjects were asked to perform well defined actions rather than in free style, because 
the authors argued that variability could bias the evaluation of the performance 
of an algorithm. Notice that all actions were captured in the same indoor 
setting and with clean background.
%\begin{figure}[htbp!]
%   \centering
%   \includegraphics[keepaspectratio=true, scale =0.35]{./images/IAS.png}
%   \caption{Sample frames of the IAS dataset~\cite{Munaro2013}.}
%   \label{fig:IAS}
%\end{figure}
\subsubsection{UCFKinect}
In order to explore the trade-off between accuracy and observational latency
when recognizing actions, UCFKinect dataset~\cite{Ellis2013}(\url{http://www.cs.ucf.edu/~smasood/datasets/UCFKinect.zip}) was created. It was 
collected by University of Central Florida Orlando in 2013. This dataset can be 
used for measuring how fast a recognition system can overcome the ambiguity in 
initial poses when performing an action. The dataset is composed of 16 actions, 
including \textit{balance, climb up, clumb ladder, duck, hop, vault, leap, run, 
kick, punch, twist left, twist right, step forward, step back, step left, step 
right}. Sixteen subjects (13 males and 3 females, all ranging between ages 20 to 
35) were involved with each subject performing all 16 actions 5 times for a 
total of 1280 action samples. The dataset is only presented as skeleton 
data comprising 3-dimensional coordinates of 15 joints along with the 
correponding orientation and binary confidence values. Subjects were 
asked to stand in a relaxed posture with loosely downward hanging arms 
beside the body before performing different actions. They were then told what 
action to perform and if requested, given a demonstration of the action. The 
end of a countdown signalled the beginning of recording and performance of the
action. The recording was manually stopped upon completion of the action. The 
authors claimed that 
gathering the data in this fashion simulates a gaming scenario where the user 
performs a variety of actions, such as punches and kicks, and returns to a 
resting pose between actions. 
%Figure~\ref{fig:UCFKinect} shows some sample skeleton frames from UCFKinect dataset.

%\begin{figure}[htbp!]
%   \centering
%   \includegraphics[keepaspectratio=true, scale =0.35]{./images/UCFKinect1.png}   
%%   \includegraphics[keepaspectratio=true, scale =0.4]{./images/UCFKinect2.png} 
%   \caption{Sample frames of skeleton data in the UCFKinect dataset~\cite{Ellis2013}.}
%   \label{fig:UCFKinect}
%\end{figure}

\subsubsection{Osaka University Kinect Action}
The Osaka University Kinect Action Dataset~\cite{Mansur2013}(\url{http://www.am.sanken.osaka-u.ac.jp/~mansur/dataset.html}) was 
collected by Osaka University in 2013 within laboratory environment. Ten actions 
were performed by 8 subjects and once. Action types are \textit{jumping jack 
type 1, jumping jack type 2, jumping on both legs, jumping on right leg, jumping 
on left leg, running, walking, side jumps, skipping on left leg,} and 
\textit{skipping on right leg}. RGB, depth, and skeleton data were all 
captured. The background and illumination conditions remained unchanged during 
the capture sessions. 
% Examples of the depth sequences are shown in Figure~\ref{fig:Osaka}.
%% \begin{figure}[htbp!]
%%    \centering
%% %    \includegraphics[keepaspectratio=true, scale =0.4]{./images/osaka.png}   
%%    \caption{Sample depth frames of Osaka dataset. From~\cite{Mansur2013}}
%%    \label{fig:Osaka}
%% \end{figure}
\subsubsection{Human Morning Routine Dataset}
Human Morning Routine dataset~\cite{Karg2013}(\url{http://www.uni-tuebingen.de/fakultaeten/mathematisch-naturwissenschaftliche-fakultaet/fachbereiche/informatik/lehrstuehle/human-computer-interaction/home/code-datasets/morning-routine-dataset.html}) was collected by Technische 
Universit\"at M\"unchen and the Eberhard Karls Universit\"at T\"ubingen in 
2013. It is aimed at testing algorithms for recognizing and monitoring morning 
routine of a human in a kitchen. A robot was supposed to be able to react to 
these activities/actions. They include \textit{preparing a 
drink}, \textit{drinking a glass of water}, \textit{preparing breakfast}, 
\textit{having breakfast}, \textit{cleaning the table}, \textit{packing a bottle 
of water into the backpack}, and \textit{leaving the room with the backpack}.  
A participant reenacted and logged his morning routine (including location he 
stood while performing those activities) in an experimental kitchen equipped 
with two Kinect~\texttrademark devices (one for motion-tracking and the other 
for detection of objects). The actions were annotated to provide ground 
truth.
% The capture environment and sample cloud can be found 
% in figure~\ref{fig:MIT}. The participant was required to write down the 
% activities that he 
% performed before going to work and the . 
%% \begin{figure}[htbp!]
%%    \centering
%% %    \includegraphics[keepaspectratio=true, scale =0.5]{./images/MITclear.png}
%%    \caption{The sample environment settings and activities, and visualization of the motion tracking data and object detections of MIT dataset. From~\cite{Karg2013}}
%%    \label{fig:MIT}
%% \end{figure}
\subsubsection{RGBD-SAR Dataset}
RGBD-SAR Dataset~\cite{Zhao2013}(\url{http://www.uestcrobot.net/en/?q=download}), created by the University of Electronic 
Science and Technology of China and Microsoft, aimed at algorithms monitoring 
behaviours of seniors. Nine categories of elderly daily activities are 
collected: 
\textit{put on 
the jacket}, \textit{take off the jacket}, \textit{enter the room}, \textit{exit 
the 
room}, \textit{sit down}, \textit{stand 
up}, \textit{drink water}, \textit{eat meal}, and \textit{walk}. 
%Figure~\ref{fig:RGBD-SAR} shows some sample frames. 
Thirty elderly 
people were invited to perform these activities and each of them performed each 
activity thrice.
%\begin{figure}[htbp!]
%   \centering
%   \includegraphics[keepaspectratio=true, scale =0.12]{./images/RGBD-SAR.jpg}
%   \caption{Sample frames of the RGBD-SAR dataset~\cite{Zhao2013}.}
%   \label{fig:RGBD-SAR}
%\end{figure}
\subsubsection{Mivia Dataset}
Mivia dataset~\cite{Carletti2013}(\url{http://mivia.unisa.it/datasets/video-analysis-datasets/mivia-action-dataset/}) was acquired by Mivia Lab at the University 
of Salemo in 2013. It consists of 7 high-level actions performed by 14 
subjects. Each subject performed 5 repetitions of each action. The actions 
include: \textit{opening a jar}, \textit{drinking}, \textit{sleeping}, 
\textit{random movements}, \textit{stopping}, \textit{interacting with 
a table} and \textit{sitting}. 
%Three sample frames are shown in Figure~\ref{fig:Mivia}.
%\begin{figure}[htbp!]
%   \centering
%   \includegraphics[keepaspectratio=true, scale =0.5]{./images/Mivia.jpg}
%   \caption{Sample frames of the Mivia dataset~\cite{Carletti2013}.}
%   \label{fig:Mivia}
%\end{figure}
\subsubsection{UPCV}
The UPCV action dataset~\cite{Theodorakopoulos2014}(\url{http://www.upcv.upatras.gr/personal/kastaniotis/datasets.html}) was collected by the 
University of Patras in 2014. The dataset consists of 10 actions performed by 
20 subjects twice. The actions, representing activities usually performed by 
ppedestrians, include: \textit{walk}, \textit{seat}, 
\textit{grab}, \textit{phone}, \textit{watch clock}, 
\textit{scratch head}, \textit{cross arms}, \textit{punch}, \textit{kick}, and 
\textit{wave}.  The published UPCV dataset only contains skeleton data. 
The subjects perform the actions in front of a fixed camera in a natural 
manner and against a stationary background. The  
ground truth provided is the annotation of data, which can isolate the action 
data from the overall motion. 
% The RGB and depth samples are shown in figure~\ref{fig:UPCV}.
%% \begin{figure}[htbp!]
%%    \centering
%% %    \includegraphics[keepaspectratio=true, scale =0.6]{./images/UPCV.jpg}
%%    \caption{Sample frames of UPCV dataset. From~\cite{Theodorakopoulos2014}}
%%    \label{fig:UPCV}
%% \end{figure}
\subsubsection{TJU dataset}
The TJU dataset~\cite{Liu2014a}(\url{http://media.tju.edu.cn/tju_dataset.html}) was captured by Tianjin University in 2014. 
and contains 22 actions performed by 20 subjects in two different 
environments; a total of 1760 sequences. Action types include: \textit{boxing, 
side boxing, one hand wave, two hands wave, hand clap, side bend, forward bend, 
draw X, draw tick, draw circle, tennis serve, tennis swing, walking, side 
walking, jogging, running, jacks, jump, jump in place, forward kick, side kick,} 
and \textit{sit down}. The 
background was fixed during capture and was subtracted
from depth data before publishing the dataset.
% The sample frames are shown in figure~\ref{fig:TJU}.
%% \begin{figure}[htbp!]
%%    \centering
%% %    \includegraphics[keepaspectratio=true, scale =0.45]{./images/TJU.png}   
%%    \caption{Sample frames of TJU dataset. From~\cite{Liu2014a}}
%%    \label{fig:TJU}
%% \end{figure}
\subsubsection{MAD}
Due to the fact that there were very few publicly available sequential action 
dataset which can be used in the development and evaluation of detection 
algorithms, the Multi-modal action detection (MAD) Dataset~\cite{Huang2014}(\url{http://humansensing.cs.cmu.edu/mad/download.html}) was 
created by Carnegie Mellon University in 2014. It contains 35 sequential actions 
performed by 20 subjects. Each subject performed the sequential actions twice. 
There are 40 sequences in total (2 sequences for each subject). The actions 
include: \textit{Running}, \textit{crouching}, \textit{jumping}, 
\textit{walking}, \textit{jump and side-kick}, \textit{left arm swipe to the 
left}, \textit{left arm 
swipe to the right}, \textit{left arm wave}, \textit{left arm punch}, 
\textit{left arm 
dribble}, \textit{left arm 
pointing to the ceiling}, \textit{left arm throw}, \textit{swing from left 
(baseball 
swing)}, \textit{left 
arm receive}, \textit{left arm back receive}, \textit{left leg kick to the 
front}, \textit{left leg kick to 
the left}, \textit{right arm swipe to the left}, \textit{right arm swipe to the 
right}, \textit{right arm 
wave}, \textit{right arm punch}, \textit{right arm dribble}, \textit{right arm 
pointing to the ceiling}, \textit{right arm throw}, \textit{swing from right 
(baseball swing)}, 
\textit{right arm 
receive, right arm back receive}, \textit{right leg kick to the front}, \textit{right leg kick to 
the right}, \textit{cross arms in the chest}, \textit{basketball shooting}, 
\textit{both arms pointing to the screen}, \textit{both 
arms pointing to both sides}, \textit{both arms pointing to right side}, 
\textit{both arms pointing to left side}. 
%Figure~\ref{fig:MAD} shows sample RGB, depth, and skeleton frames of three actions.
The authors provided ground truth labels which 
indicated the start and end of the actions and are suitable for both detection 
and classification.
%\begin{figure}[htbp!]
%   \centering
%   \includegraphics[keepaspectratio=true, scale =0.32]{./images/MAD.jpg}
%   \caption{Sample frames of the MAD dataset~\cite{Huang2014}.}
%   \label{fig:MAD}
%\end{figure}
\subsubsection{Composable activities}
Composable activities dataset~\cite{Lillo2014}(\url{http://web.ing.puc.cl/~ialillo/ActionsCVPR2014/}) was created by Pontificia 
Universidad Catolica de Chile and Universidad del Norte in 2014. It was aimed 
at the problem of recognizing complex activities, such as \textit{waving while 
walking}, \textit{talking on the phone while running away to attend an urgent 
matter}, etc. 
Different combinations of 26 atomic actions formed 16 activity classes which 
were performed by 14 subjects and annotations were provided. Each activity is 
composed of 3 to 11 atomic 
actions. For example, the activity \textit{walk while hand waving} consists of 3 
atomic actions: \textit{walk}, \textit{hand wave}, and \textit{idle}; while the 
activity \textit{composed-activity-4} is composed of 11 atomic actions: 
\textit{idle}, \textit{walk}, \textit{call a friend with hands}, \textit{hand 
wave}, 
\textit{talking on cellphone}, \textit{pick from the floor}, \textit{dial 
cellphone}, \textit{put an object}, \textit{pick 
cellphone from pocket}, and \textit{put cellphone in pocket}. 
\subsubsection{3D Online Action}
3D online action dataset~\cite{Yu2015}(\url{https://sites.google.com/site/skicyyu/rgbd_recognition}) was collected by  Microsoft and Nanyang 
Technological University in 2014 with the aim of developing and 
testing algorithms for continuous online human action recognition from RGB-D 
data. There are seven action categories: 
\textit{drinking}, 
\textit{eating}, \textit{using laptop}, \textit{reading cellphone}, 
\textit{making phone 
call}, \textit{reading book} and \textit{using 
remote}. 
%Sample frames from depth and skeleton videos are shown in Figure~\ref{fig:ORGBD}. 
Thirty-six subjects performed the actions in this dataset. 
The dataset is intended for the evaluation of three categories of 
tasks: same-environment action recognition, 
cross-environment action recognition, and continuous action recognition. In 
order to achieve this purpose, the dataset was separated into four sections: 
first two sections contain single action in each sample and were captured in 
same environment; the third section also contains single action in each sample, 
but was captured in a different environment; the fourth section contains 
multiple, albeit orderless actions in each sample. The bounding box of the 
object involved in each frame is manually labelled.
%\begin{figure}[htbp!]
%   \centering
%   \includegraphics[keepaspectratio=true, scale =0.35]{./images/ORGBD.png}
%   \caption{Sample frames of the 3D Online Action dataset~\cite{Yu2015}.}
%   \label{fig:ORGBD}
%\end{figure}
\subsubsection{RGB-D activity dataset}
The RGB-D activity dataset~\cite{Wu2015}(\url{http://watchnpatch.cs.cornell.edu/}) was collected by Cornell
University and Stanford University in 2015. The dataset was recorded by the 
Kinect v2 camera. Each video in the dataset contains 2-7 actions involving 
interaction with different objects. Compared to previous Kinect v1 system, the 
Kinect v2 has higher resolution of RGB-D data (RGB: 1920*1080, depth: 512*424) 
and improved body tracking of human skeletons (25 body joints). In this dataset, 
21 actions (10 in the office, 11 in the kitchen) interacted with 23 types of 
objects were performed by 7 subjects. The action  categories are: 
\textit{turn-on-monitor, turn-off-monitor, walking, play-computer, reading, 
fetch-book, put-back-book, take-item, put-down-item, leave-office, 
fetch-from-fridge, put-back-to-fridge, prepare-food, microwaving, 
fetch-from-oven, pouring, drinking, leave-kitchen, move-kettle, fill-kettle, and 
plug-in-kettle}. The background of the captured scene are relatively complex 
and in each environment the activities were performed relative to different 
views. In total, there are 458 videos with a total length of about 230 minutes. 
%Figure~\ref{fig:RGB-DActivity} shows some sample frames.

%\begin{figure}[htbp]
%\centering
%\subfloat[turn-off-monitor]{
%	\begin{minipage}[t]{0.5\textwidth}
%		\centering
%		\includegraphics[keepaspectratio=true, scale =0.25]{./images/RGBD-Activity1.jpg}
%	\end{minipage}
%	}
%	\subfloat[take-item]{
%	\begin{minipage}[t]{0.5\textwidth}
%		\centering
%		\includegraphics[keepaspectratio=true, scale =0.25]		{./images/RGBD-Activity2.jpg}
%	\end{minipage}
%	}
%	\caption{Sample frames of the RGB-D activity dataset~\cite{Wu2015}.}
%    \label{fig:RGB-DActivity}
%\end{figure}

\subsubsection{SYSU 3D Human-Object Interaction Dataset}
The SYSU 3D Human-Object Interaction dataset~\cite{Hu2015}(\url{http://sist.sysu.edu.cn/~zhwshi/students/jianfang/HomePage.htm}) was created by Sun 
Yat-sen University in 2015. This dataset focuses on actions involving 
human-object interaction. Forty subjects perform 12 distinct activities, such 
as \textit{drinking, pouring, calling phone, 
playing phone, wearing backpacks, packing backpacks, sitting chair, moving 
chair, taking out wallet, taking from wallet, mopping}, and \textit{sweeping}. 
For each activity, each subject manipulates one of the six different objects: 
phone, chair, bag, wallet, mop and besom. Hence, the dataset contains 480 video 
clips in total. The RGB frames, depth sequence and skeleton data of each video 
clips are captured by a Kinect camera. The authors claimed that their dataset 
presents some new challenges compared to previous datasets. For example,  
the  motions and the appearance of manipulated objects are highly similar 
between some activities, and the number of participants is larger than that of 
any existing dataset. 
% Figure~\ref{fig:SYSU} gives some sample frames.
% 
%% \begin{figure}[htbp!]
%%    \centering
%% %    \includegraphics[keepaspectratio=true, scale =0.7]{./images/SYSU.png}   
%%    \caption{Sample frames of SYSU 3D Human-Object Interaction dataset. From~\cite{Hu2015}}
%%    \label{fig:SYSU}
%% \end{figure}
\subsubsection{UTD-MHAD}
UTD-MHAD~\cite{Chen2015b}(\url{http://www.utdallas.edu/~cxc123730/UTD-MHAD.html}) was collected by University of Texas at Dallas in 2015. Eight subjects performed 27 actions four times. The 27 actions are: \textit{right arm swipe to the left, right arm swipe to the right, right hand wave, two hand front clap, right arm throw, cross arms in the chest, basketball shoot, right hand draw x, right hand draw circle (clockwise), right hand draw circle (counter clockwise), draw triangle, bowling (right hand), front boxing, baseball swing from right, tennis right hand forehand swing, arm curl (two arms), tennis serve, two hand push, right hand knock on door, right hand catch an object, right hand pick up and throw, jogging in place, walking in place, sit to stand, stand to sit, forward lunge (left foot forward)}, and \textit{squat (two arms stretch out)}. All the actions were performed in a fixed background. An inertial sensor was worn on the subject's right wrist for action 1 to 21, and on the right thigh for action 22 to 27. Hence, four types of data modalities were captured, namely RGB videos, depth videos, skeleton joint positions, and the inertial sensor signals.

\begin{scriptsize}
\begin{longtable}[HTBP]{|p{3.3cm}|p{1.5cm}|P{1.3cm}|P{1.5cm}|P{6.8cm}|}

\hline
Dataset & Year(Cited\footnote{Citations as of 31 August 2015}) & Modality & \#a,\#s,\#e & Protocol\\
\hline\hline

MSR-Action3D ~\cite{Li2010} &  2010 (333) & D,S &  20,10,567 & \setlist{nolistsep,leftmargin=*} \begin{enumerate}[topsep=0pt, partopsep=0pt, 
itemsep=0pt,parsep=0pt]
      \item 1/3 training
      \item 2/3 training
      \item Half training, half testing CS \vspace{-1em}
\end{enumerate} \\
\hline
MSRDaily-Activity3D~\cite{Wang2012} & 2012 (311) & C,D,S & 16,10,320 & Half training, half test CS \\
\hline
UTKinect ~\cite{Xia2012} & 2012,(193) & C,D,S & 10,10,200 
& LOSeqO\\
\hline
CAD-60 ~\cite{Sung2011} & 2011(159) & C,D,S & 12,4,60 
& \setlist{nolistsep,leftmargin=*} \begin{enumerate}[topsep=0pt, partopsep=0pt, 
itemsep=0pt,parsep=0pt]
      \item LOSubO
      \item Halved the testing subject's data and included one half in the training dataset \vspace{-1em}
\end{enumerate} \\
\hline
RGBD-HuDaAct ~\cite{Ni2011} & 2011(148) & C,D & 12,30,1189 
& LOSubO\\
\hline
MSRAction-Pair ~\cite{Oreifej2013} & 2013(136) & C,D,S & 12,10,180  
& First half training CS \\
\hline
MSRC-12 gesture ~\cite{Fothergill2012} & 2012(100) & S & 12,30,594 & LOSubO \\
\hline
CAD-120 ~\cite{Koppula2013a} & 2013(81)& C,D,S & 10,4,120 & 
LOSubO (4-fold CV)\\
\hline
UCFKinect ~\cite{Ellis2013} & 2013(62) & S & 16,16,1280 & 4-fold CV \\
\hline 
G3D ~\cite{Bloom2012, Bloom2013} & 2012(28) & C,D,S & 20,10,234 & CS (4 subjects training, 1 subject validation, 5 subjects test) \\
\hline
Falling Event ~\cite{Zhang2012a} & 2012(21) & C,D,S & 5,5,150 & 50 samples covering 5 subjects and 5 activities with sufficient lighting for training, rest for testing \\
\hline 
UPCV ~\cite{Theodorakopoulos2014} & 2014(18) & S & 10,20,400 & 
LOSubO\\
\hline
DHA ~\cite{Lin2012} & 2012(17) & C,HM,D & 23,21,483 &  
 CS (10 training,11 test) \\
\hline
WorkoutSU-10 ~\cite{Negin2013} & 2013(16) & C,D,S & 10,12,1200 & CS\\
\hline 
IAS-lab ~\cite{Munaro2013, Munaro2013a} & 2013(15) & C,D,S,P & 15,12,540 & LOSubO\\
\hline
Osaka ~\cite{Mansur2013} & 2013(8) & C,D,S & 10,8,80 & LOSubO CV \\
\hline 
Mivia ~\cite{Carletti2013} & 2013(6) & C,D & 7,14,490 &
\setlist{nolistsep,leftmargin=*} \begin{enumerate}[topsep=0pt, partopsep=0pt, 
itemsep=0pt,parsep=0pt]
      \item Leave two repetitions of one person out. 
      \item LOSubO \vspace{-1em}
\end{enumerate} \\
\hline
Concurrent Action~\cite{Wei2013a} & 2013(5) & S & 12,-,61 & 
Not given \\
\hline
TJU ~\cite{Liu2014a} & 2014(4) & C,D,S & 22,20,1760 & First 12 subjects training, rest test \\
\hline 
3D Online ~\cite{Yu2015} & 2014(4) & C,D,S & 7,36,386 & 
\setlist{nolistsep,leftmargin=*} \begin{enumerate}[topsep=0pt, partopsep=0pt, itemsep=0pt,parsep=0pt]
      \item Same-Environment (2-fold CV)
      \item Cross-Environment (S1, S2 training, S3 test)
      \item Continuous (S1, S2, S3 training, S4 test) \vspace{-1em}
\end{enumerate} \\
\hline
MAD ~\cite{Huang2014} & 2014(3) & C,D,S & 20,35,40 & 
 5-fold CV (8 groups training, 2 groups test)\\
\hline
Composable ~\cite{Lillo2014} & 2014(3) & C,D,S & 16,14,693 & LOSubO \\
\hline
RGBD-SAR ~\cite{Zhao2013} & 2013(1) & C,D & 12,6,810 & 
Not given\\
\hline
SYSU ~\cite{Hu2015} & 2015(0) & C,D,S & 12,40,480 & \setlist{nolistsep,leftmargin=*} \begin{enumerate}[topsep=0pt, partopsep=0pt, 
itemsep=0pt,parsep=0pt]
      \item Half samples training, rest test
      \item CS \vspace{-1em}
\end{enumerate}\\
\hline 
RGB-D activity ~\cite{Wu2015} & 2015(0) & C,D,S & 21,7,458 & Not given \\
\hline 
UTD-MHAD ~\cite{Chen2015b} & 2015(0) & C,D,S,I & 27,8,861 & CS (odd subjects training, even subjects test) \\
\hline
Morning-Routine~\cite{Karg2013} & 2013(0) & C,D,S & 7,1,14 & Not given \\
\hline 

\captionsetup{width=\textwidth, belowskip=10pt,aboveskip=6pt}
\caption{Summary of basic specifications of Single-view action/activity 
datasets. Notation for the header: \#a: number of actions, \#s: number of 
subjects, \#e: number of total examples. Notation for data format: C: Colour, D: 
Depth, S: Skeleton, HM: Human Mask, P: Point clouds, I: Inertial sensor data. Notation for protocol: CS: 
Cross Subject, LOSeqO: Leave One Sequence Out, LOSubO: Leave One Subject Out, 
CV: Cross Validation}
\label{table:Basicsingle}
\end{longtable}
\end{scriptsize}

\subsection{Multi-view action/activity datasets}
\label{sec:mutiview}
A multi-view dataset can be generated in at least two ways. First, several 
cameras can be mounted at different positions and angles. Second, the same 
action can be repeated from different viewpoints. 
The reviewed multiview datasets are generated using these two approaches. 
However, most of them are captured by multiple cameras. Similarly to the 
review of single-view datasets, the descriptions of multiview datasets are 
given in chronological order. Table~\ref{table:Basicmultiview} shows a summary 
of basic specifications of multi-view datasets.

% Multi-view can be obtained from two aspects. First is mounting more than one 
% cameras from different directions or angles. Second is repeating a same action 
% from various view points. As a result, the multi-view datasets are generally 
% generated according to above two approaches. 

\subsubsection{ATC$4^2$}
ATC$4^2$ dataset~\cite{Cheng2012}(\url{http://vipl.ict.ac.cn/rgbd-action-dataset})was collected by Institute of Computing 
Technology of Chinese Academy of Science in 2012 for the purpose of providing  
an evaluative framework that supports view variations of actions. The dataset 
focuses on facilitating practical applications, such as smart house or 
e-healthcare, 
and contains 14 daily activities: \textit{Collapse}, \textit{Drink}, 
\textit{MakePhonecall}, \textit{MopFloor}, 
\textit{PickUp}, \textit{PutOn}, \textit{ReadBook}, \textit{SitDown}, 
\textit{SitUp}, \textit{Stumble}, \textit{TakeOff}, \textit{ThrowAway}, 
\textit{TwistOpen}, \textit{WipeClean}. Note that \textit{Collapse} and 
\textit{Stumble} are two activities specific to homecare applications. The 
authors distinguished between \textit{Collapse} (people falling as a result of 
inner factors, such as hurt or giddiness) and \textit{Stumble} (body 
dropping caused by outside effects such as tripping on an obstacle). The 
dataset was captured by 4 Kinect sensors from different heights and view 
angles. 
%The environment setting and one sample depth frame are shown in Figure~\ref{fig:ATC$4^2$}. 
Twenty-four subjects performed the 14 activities for several times. 
The labels of start/stop points of single actions are provided.
%\begin{figure}[htbp!]
%   \centering
%   \includegraphics[keepaspectratio=true, scale =0.28]{./images/ATC42clear2.png}
%   \caption{Environment setting and one sample frame of the ATC$4^2$ dataset~\cite{Cheng2012}.}
%   \label{fig:ATC$4^2$}
%\end{figure}
\subsubsection{Falling Detection}
The Falling Detection dataset~\cite{Zhang2012b}(\url{http://vlm1.uta.edu/~zhangzhong/fall_detection/}) was collected by the University 
of Texas in 2012. It focused on falling actions captured 
in a laboratory-based simulated apartment set up. Six subjects in two sceneries 
performed a series of actions continuously, including both \textit{real fall 
actions} and fall-like actions, such as \textit{picking up a coin from floor, 
sitting down on the floor, tying shoelaces, sleeping down on the bed, opening 
the lower drawer which is close to the floor, jumping on to the floor,}
\textit{and sleeping down on the floor}. Only depth data sequences
are published along with annotation of the start and end 
frame for every fall process, but not other actions. There are 12 real falls in 
video from the first scene, and 14 real falls 
in the second scene. For the fall like actions, there are 23 examples
of picking up something from the floor, 12 cases
of sitting on the floor, 10 examples of tying shoelaces,
9 examples of lying down on the bed, 5 examples of
opening/closing a drawer at floor level, 1 example of
jumping on the bed, and 1 example of lying on the floor. 
%Figure~\ref{fig:FallDetection} shows the sample frames of a falling action.

%\begin{figure}[htbp!]
%   \centering
%   \includegraphics[keepaspectratio=true, scale =0.3]{./images/FallingDetection.png}   
%   \caption{Sample frames of the Falling Detection dataset~\cite{Zhanga}}
%   \label{fig:FallDetection}
%\end{figure}
\subsubsection{Berkeley MHAD}
Berkeley Multimodal Human Action Database (MHAD) ~\cite{Ofli2013}(\url{http://tele-immersion.citris-uc.org/berkeley_mhad#dl}), collected by 
University of California at Berkeley and Johns Hopkins University in 2013, was captured 
in five different modalities to expand the fields of application. The 
modalities are derived from: optical mocap system, four multi-view stereo vision 
cameras, two Microsoft Kinect~\texttrademark cameras, six wireless 
accelerometers and four microphones. Twelve subjects performed 11 actions, 
five times each. Three categories of actions are included: (1) actions with 
movement in full body 
parts, e.g., \textit{jumping in place}, \textit{jumping jacks}, 
\textit{throwing}, etc., (2) actions with high dynamics in upper 
extremities, e.g.,\textit{ waving hands}, \textit{clapping hands}, etc. and (3) 
actions with high dynamics in lower extremities, e.g., \textit{sit down}, 
\textit{stand 
up}. The actions were executed with style and speed variations. This dataset 
can be used in the development and evaluation of multimodal algorithms, 
such as action recognition, pose estimation, motion segmentation and dynamic 3D 
scene reconstruction.
%Sample images and point clouds of each action are shown in Figure~\ref{fig:BerkeleyMHAD}. 

%\begin{figure}[htbp!]
%   \centering
%   \includegraphics[keepaspectratio=true, scale =0.2]{./images/BerkeleyMHADclaer.jpg}
%   \caption{Sample images and point clouds of the BerkeleyMHAD dataset~\cite{Ofli2013}.}
%   \label{fig:BerkeleyMHAD}
%\end{figure}
\subsubsection{DMLSmartActions}
DMLSmartActions dataset~\cite{Amiri2013}(\url{http://dml.ece.ubc.ca/data/smartaction/}) was collected by the University of 
British Columbia in 2013 and aimed at demonstrating the 
real situation in a home environment. 
Two high-definition (HD) RGB cameras and one Kinect sensor were utilized for collecting 
the data. Although the three cameras were static during acquisition, their 
location and orientation were not fixed so as to provide variability. 
The Kinect~\texttrademark sensor was always located between the two HD RGB 
cameras in different scenes. Sixteen subjects performed 12 different actions in 
a natural manner. The actions include: \textit{clean-table}, \textit{drink}, 
\textit{drop-and-pickup}, \textit{fell-down}, \textit{pick-something}, 
\textit{put-something}, \textit{read}, \textit{sit-down}, \textit{standup}, 
\textit{use-cellphone}, \textit{walk}, and \textit{write}. Subjects were 
asked to perform a series of the listed actions in a natural style, 
suggesting that there was no instruction on how or when to perform these 
actions. 
%Figure~\ref{fig:DMLSmartActions} shows several sample frames of each modality. 
The data was manually labelled into samples.

%\begin{figure}[htbp!]
%\begin{center}
%\begin{tabular}{m{3.8cm}<{\centering}m{3.8cm}<{\centering}m{2.8cm}<{\centering}m{3cm}<{\centering}}
%HD-1 & HD-1 & Kinect-Depth & Kinect-RGB \\
%
%	\begin{minipage}{.3\textwidth}
%      \includegraphics[keepaspectratio=true, scale =0.35]{./images/DMLSmart1.jpg}
%    \end{minipage} & 
%    \begin{minipage}{.3\textwidth}
%      \includegraphics[keepaspectratio=true, scale =0.35]{./images/DMLSmart2.jpg}
%    \end{minipage} &
%    \begin{minipage}{.3\textwidth}
%      \includegraphics[keepaspectratio=true, scale =0.35]{./images/DMLSmart3.jpg}
%    \end{minipage} & 
%    \begin{minipage}{.3\textwidth}
%      \includegraphics[keepaspectratio=true, scale =0.35]{./images/DMLSmart4.jpg}
%    \end{minipage} 
%\\
%\end{tabular}
%\caption{Sample frames of the DMLSmartActions dataset~\cite{Amiri2013}.}
%   \label{fig:DMLSmartActions}
%\end{center}
%\end{figure}

\subsubsection{ReadingAct}
ReadingAct dataset~\cite{Chen2013} was collected by Reading University in 2013, using 2 Kinect sensors; one was in front of the 
subject and the other was placed orthogonally to capture a side view. Twenty 
actors performed the actions four times in free form style to ensure variability. The 
dataset includes a background scene and 19 actions: \textit{coming in}, 
\textit{going out}, \textit{walking past}, \textit{walking around}, 
\textit{switching light}, \textit{talking on phone}, \textit{phone 
call (mobile)}, \textit{picking up from floor}, \textit{putting on jacket}, 
\textit{hoovering floor}, \textit{sitting down}, \textit{standing up}, 
\textit{lying down}, \textit{getting up}, \textit{reading a book}, 
\textit{typing on computer}, \textit{having meal}, \textit{drinking 
(sitting)} and \textit{drinking (standing)}. 
%Sample frames of ReadingAct dataset can be found in Figure~\ref{fig:ReadingAct}.
%\begin{figure}[htbp!]
%   \centering
%   \includegraphics[keepaspectratio=true, scale =0.3]{./images/ReadingAct.png}
%   \caption{Sample frames of the ReadingAct dataset~\cite{Chen2013}.}
%   \label{fig:ReadingAct}
%\end{figure}
\subsubsection{Multiview 3D Event}
Multiview 3D Event dataset~\cite{Wei2013}(\url{http://www.stat.ucla.edu/~ping.wei/research/project/4DHOI/4DHOI.html}) was created by University of 
California at Los Angles in 2013 using three simultaneous Kinect~\texttrademark 
sensors from different viewpoints around the subjects. 
%Sample frames captured from three view points are shown in Figure~\ref{fig:Multiview3DEvent}. 
This dataset includes 8 categories of events performed by 8 subjects 20 times 
independently with different object instances and in various styles. The eight 
event categories are: \textit{drink with mug}, \textit{call with cellphone}, 
\textit{read book}, \textit{use mouse}, \textit{type on keyboard}, 
\textit{fetch water from dispenser}, \textit{pour water from kettle}, and 
\textit{press button}. These events involve 11 object classes: mug, cellphone, 
book, mouse, keyboard, dispenser, kettle, button, monitor, chair, and desk. To 
label the data, the videos were manually cut into sequences wherein  each 
sequence contains one action.
%The datasets can be used in event recognition, sequence segmentation, 
% object recognition and localization as both the action labels and object labels 
% are provided.

%\begin{figure}[htbp!]
%\begin{center}
%\begin{tabular}{m{4cm}<{\centering}m{4cm}<{\centering}m{4cm}<{\centering}}
%Viewpoint 1 & Viewpoint 2 & Viewpoint 3\\
%
%	\begin{minipage}{.3\textwidth}
%      \includegraphics[keepaspectratio=true, scale =0.15]{./images/Multiview3DEvent1.jpg}
%    \end{minipage} & 
%    \begin{minipage}{.3\textwidth}
%      \includegraphics[keepaspectratio=true, scale =0.15]{./images/Multiview3DEvent2.jpg}
%    \end{minipage} &
%    \begin{minipage}{.3\textwidth}
%      \includegraphics[keepaspectratio=true, scale =0.15]{./images/Multiview3DEvent3.jpg}
%    \end{minipage} 
%\\
%\end{tabular}
%\caption{Sample images of three views of Multiview 3D Event dataset. From~\cite{Wei2013}}
%   \label{fig:Multiview3DEvent}
%\end{center}
%\end{figure}

\subsubsection{Northwestern-UCLA Multiview Action 3D}
Northwestern-UCLA Multiview Action 3D~\cite{Wang2014}(\url{http://users.eecs.northwestern.edu/~jwa368/my_data.html}) was collected by 
Northwestern University and  University of California at Los Angles in 2014. 
The capture settings were similar to Multiview 3D Event dataset but adds 
multiple locations. The actions were performed by 10 actors and captured by 
three simultaneous Kinect cameras. There are 10 action categories: 
\textit{pick up with one hand}, \textit{pick up with two hands}, \textit{drop 
trash}, \textit{walk around}, \textit{sit 
down}, \textit{stand up}, \textit{donning}, \textit{doffing}, \textit{throw}, 
\textit{carry}. 
%Sample frames of the cross environment data can be found in Figure~\ref{fig:NorthwesternUCLAMultiview}. 
%% Ten actors performed 
%% these actions. 
%% % The original application of 
%% this dataset is to evaluate a newly proposed generative cross-view video action 
%% representation.
%\begin{figure}[htbp!]
%   \centering
%   \includegraphics[keepaspectratio=true, scale =0.3]{./images/UCLA.jpg}
%   \caption{Sample frames of cross environment data of the Northwestern UCLA Multiview dataset~\cite{Wang2014}.}
%   \label{fig:NorthwesternUCLAMultiview}
%\end{figure}
\subsubsection{UWA3D Multiview}
UWA3D Multiview Activity Dataset~\cite{Rahmani2014}(\url{http://staffhome.ecm.uwa.edu.au/~00053650/databases.html}) was collected by the 
University of Western Australia in 2014. In this dataset, all actions were 
captured continuously without break or pause. Thirty activities were performed 
by 10 individuals: \textit{one hand waving}, \textit{one hand Punching}, 
\textit{sitting 
down}, \textit{standing 
up}, \textit{holding chest}, \textit{holding head}, \textit{holding back}, 
\textit{walking}, \textit{turning around}, 
\textit{drinking}, \textit{bending}, \textit{running}, \textit{kicking}, 
\textit{jumping}, \textit{moping floor}, \textit{sneezing}, \textit{sitting 
down(chair)}, \textit{squatting}, \textit{two hand waving}, \textit{two hand 
punching}, \textit{vibrating}, \textit{falling 
down}, \textit{irregular walking}, \textit{lying down}, \textit{phone 
answering}, \textit{jumping jack}, \textit{picking up}, 
\textit{putting down}, \textit{dancing}, and \textit{coughing}. Each subject 
performed the 30 activities twice or thrice continuously in random order. To 
achieve multiview, five subjects performed 15 activities from four different 
side views. 
% The sample point clouds of different view are shown in 
% figure~\ref{fig:UWA3D}. 
%% \begin{figure}[htbp!]
%%    \centering
%% %    \includegraphics[keepaspectratio=true, scale =0.5]{./images/UWA3Dpoint.png}
%%    \caption{Sample point clouds of different view of UWA3D dataset. From~\cite{Rahmani2014}}
%%    \label{fig:UWA3D}
%% \end{figure}

\subsubsection{Muti-View TJU dataset}
The Muti-View TJU dataset~\cite{Liu2015a}(\url{http://media.tju.edu.cn/tju_dataset.html}) was captured by Tianjin University in 
2014 and represents similar action types as in TJU dataset. However, this 
dataset was captured with two Kinect cameras from two viewpoints (front view 
and side view) and the angle between the two views is around 65 degrees. The 22 
actions were performed by 20 subjects four times in both light and dark 
environments. There are 7040 samples in total. Each action was recorded in 
modes RGB, depth, skeleton data, and human mask. 
%Figure~\ref{fig:MVTJU} shows sample frames of RGB and depth data from two views.
%\begin{figure}[htbp!]
%   \centering
%   \includegraphics[keepaspectratio=true, scale =0.35]{./images/TJUmulti.jpg}   
%   \caption{Sample frames of the Multi-View TJU dataset~\cite{Liu2015a}.}
%   \label{fig:MVTJU}
%\end{figure}
\subsubsection{NJUST RGB-D Action}

NJUST RGB-D Action dataset~\cite{Song2014}(\url{http://imag.njust.edu.cn/imag/NJUST_RGB-D_Action_Dataset.html}
) was collected by Nanjing 
University of Science and Technology in 2014. The dataset was collected in lab 
environments with subjects located at about three meters from the camera. 
There are 19 action categories: \textit{Bending, Bending-side, Boxing, 
Checking-Time, 
Drinking, DroppingBag, Kicking, LyingDown, OpeningCloset, PickingUp, PullingOut, 
SittingDown, Squatting, StandingUp, TakingPhoto, Telephoning, Tossing, Walking}, 
and \textit{Waving}. Each action was performed by ten subjects in two scenes. 
This dataset also provides some view variation samples of six actions. To achieve 
view variation, the subjects were asked to perform the six actions with 30 
degree view angle to the camera. The six actions are: \textit{Bending-30D, 
Boxing-30D, Drinkin-30D, SittingDown-30D, Squatting-30D and StandingUp-30D}. 
Altogether, there are 500 action samples. For each sample, RGB frames, depth 
frames, skeleton data, and body segmentation are provided. 
% The frame rate is 30 
% frames per second and the resolution is $320\times240$ pixels.
% Several samples 
% are shown in Figure~\ref{fig:NJUST}.
% 
%%\begin{figure}[htbp!]
%%   \centering
%%   \includegraphics[keepaspectratio=true, scale =0.5]{./images/NJUST.png}   
%%   \caption{Sample frames of NJUST dataset. From~\cite{Song2014}}
%%   \label{fig:NJUST}
%%\end{figure}

\newcommand{\tabincell}[2]{\begin{tabular}{@{}#1@{}}#2\end{tabular}}
\begin{scriptsize}
\begin{longtable}[HTBP]{|p{3cm}|p{1.5cm}|P{1.8cm}|P{1.5cm}|P{6cm}|}
\hline
Dataset & Year(Cited\footnote{Citations are as of 31 August 2015}) & Modality & \#a,\#s,\#e & Protocol\\
\hline\hline
Berkeley MHAD ~\cite{Ofli2013} & 2013(50) & \tabincell{c}{C, D, M, A, Au} & 12,12,720 & CS (First 7 training, last 5 test) \\
\hline
ATC$4^2$ ~\cite{Cheng2012} & 2012(27) & C,D & 14,24,6844 & 
8 training, 16 test CS \\
\hline
Falling Detection ~\cite{Zhang2012b} & 2012(17) & D & 8,6,12 & CS \\
\hline
Multiview 3D Event ~\cite{Wei2013} & 2013(13) & C,D,S & 8,8,3815 & Not given \\
\hline
Multi-View TJU ~\cite{Liu2015a} & 2013(6) & C,D,S & 20,22,7040 & 6 subjects training, 6 validation, 8 test \\
\hline
Northwestern-UCLA ~\cite{Wang2014} & 2014(5) & C,D,S & 10,10,- & 
\setlist{nolistsep,leftmargin=*} \begin{enumerate}[topsep=0pt, partopsep=0pt, 
itemsep=0pt,parsep=0pt]
      \item LOSubO
      \item 2 Camera training,1 Camera test
      \item test on different environment \vspace{-1em}
\end{enumerate} \\
\hline
UWA3D Multiview ~\cite{Rahmani2014} & 2014(4) & D,S & 30,10,600+ & 
\setlist{nolistsep,leftmargin=*} \begin{enumerate}[topsep=0pt, partopsep=0pt, 
itemsep=0pt,parsep=0pt]
      \item CS (Half training, half test) 
      \item $0^{\circ}$ training \vspace{-1em}
\end{enumerate} \\
\hline
NJUST ~\cite{Amiri2013} & 2014(2) & C,D,S,HM & 19,10,500 & LOSubO CV\\
\hline
DMLSmart Actions ~\cite{Amiri2013} & 2013(2) & HDC,C,D & 12,16,932 & LOSubO\\
\hline
ReadingAct ~\cite{Chen2013} & 2013(1) & C,D & 19,20,2340 & 
CS (15 training, 5 test, 4-fold CV) \\
\hline
\captionsetup{width=\textwidth}
\caption{Summary of basic specifications of Multi-view action/activity datasets. Notation for the header: \#a: number of actions, \#s: number of subjects, \#e: number of total examples. Notation for data format: C: Colour, D: Depth, S: Skeleton, M: Mocap, SV: Stereo Video, Au: Acceleration, A: Audio, HM: Human Masks, HDC: High Definition Colour. Notation for protocol: CS: Cross Subject, LOSubO: Leave One Subject Out, CV: Cross Validation}
\label{table:Basicmultiview}
\end{longtable}
\end{scriptsize}

\subsection{Interaction/Multi-person activity datasets}
%In this section, the focus is on datasets with actions involving more
%than one person, including human-human interaction and multi-person activity. 
The human-human interaction datasets normally contain interaction between two 
persons. The number of persons involved in multi-person activity is not 
fixed. A summary of basic specifications of interaction/multi-person activity 
datasets is provided in Table~\ref{table:Basicmultiperson}.

\subsubsection{SBU Kinect Interaction Dataset}
SBU~\cite{Yun2012}(\url{http://www3.cs.stonybrook.edu/~kyun/research/kinect_interaction/index.html}
) was collected by Stony Brook University in 2012. It contains 
eight types of interactions, including: \textit{approaching}, 
\textit{departing}, \textit{pushing}, 
\textit{kicking}, \textit{punching}, \textit{exchanging objects}, 
\textit{hugging}, and \textit{shaking hands}. 
%The sample RGB, depth, and skeleton frames of each action are shown in Figure~\ref{fig:SBU}. 
All videos were recorded with the same indoor background. Seven participants were 
involved in performing the activities which have interactions 
between two actors. The dataset is segmented into 21 sets and each set contains 
one or two sequences of each action category. Two kinds of ground truth 
information are provided: action labels of each segmented video and 
identification of ``active'' actor and ``inactive'' actor.

%\begin{figure}[htbp]
%	\centering
%	\subfloat[Approaching]{
%	\begin{minipage}[t]{0.23\textwidth}
%		\centering
%		\includegraphics[keepaspectratio=true, scale =0.28]{./images/SBU1.png}
%	\end{minipage}
%	}
%	\subfloat[Departing]{
%	\begin{minipage}[t]{0.23\textwidth}
%		\centering
%		\includegraphics[keepaspectratio=true, scale =0.28]{./images/SBU2.png}
%	\end{minipage}
%	}
%	\subfloat[Kicking]{
%	\begin{minipage}[t]{0.23\textwidth}
%		\centering
%		\includegraphics[keepaspectratio=true, scale =0.28]{./images/SBU3.png}
%	\end{minipage}
%	}
%	\subfloat[Punching]{
%	\begin{minipage}[t]{0.23\textwidth}
%		\centering
%		\includegraphics[keepaspectratio=true, scale =0.28]{./images/SBU4.png}
%	\end{minipage}
%	}
%	\caption{Sample frames of the SBU dataset~\cite{Yun2012}.}
%    \label{fig:SBU}
%\end{figure}

\subsubsection{K3HI}
Similarly to SBU dataset, K3HI~\cite{Hu2013}(\url{http://www.lmars.whu.edu.cn:8086/prof_web/zhuxinyan/DataSetPublish/dataset.html}
) is also a two-person interaction 
dataset. It was collected by Wuhan University in 2013. Fifteen volunteers performed 8 
categories of activities, including \textit{approaching}, \textit{departing}, 
\textit{kicking}, \textit{punching}, 
\textit{pointing}, \textit{pushing}, \textit{exchanging an object}, and 
\textit{shaking 
hands}.
% The pair of actors performed all types of interactions.  
In order to ensure the integrity and continuity of the spatial information of 
the skeleton data of the two persons, the RGB and depth data were ignored 
during data capture. 
%% Figure~\ref{fig:K3HI} shows the sample skeleton of six actions. 
%% \begin{figure}[htbp!]
%%    \centering
%% %    \includegraphics[keepaspectratio=true, scale =0.38]{./images/K3HIclear.png}
%%    \caption{Sample frames of K3HI dataset. From~\cite{Hu2013}}
%%    \label{fig:K3HI}
%% \end{figure}
\subsubsection{The LIRIS human activities dataset}
LIRIS Human Activities Dataset~\cite{Wolf2014}(\url{http://liris.cnrs.fr/voir/activities-dataset/}), collected by the French National Center for Scientific Research in 2014, was captured in complex 
scenarios. The Kinect~\texttrademark sensor was mounted on a remotely controlled 
robot to capture activities involving human-human interactions, 
human-object interactions and human-human-object interactions. All the 
activities were examples from daily life, such as \textit{discussing}, 
\textit{telephone calls}, 
\textit{giving an item}, etc. 
%(see Figure~\ref{fig:LIRIS} ). 
Full localization information with bounding boxes is provided as ground truth for each frame of each activity. 
%\begin{figure}[htbp!]
%   \centering
%   \includegraphics[keepaspectratio=true, scale =0.3]{./images/LIRIS.jpg}
%   \caption{Sample frames of the LIRIS dataset~\cite{Wolf2014}.}
%   \label{fig:LIRIS}
%\end{figure}
\subsubsection{G3Di}
G3Di~\cite{Bloom2014}(\url{http://dipersec.king.ac.uk/G3D/}) is a human interaction dataset for multiplayer gaming 
scenarios and was collected by the same group that colected G3D dataset at 
Kingston University in 2014. The dataset was captured using a gamesourcing approach 
where the users were recorded whilst playing computer games. This dataset 
contains 12 subjects split into 6 
pairs. Each pair interacted through a gaming interface showcasing six sports 
involving several actions: boxing (\textit{right punch}, \textit{left punch}, 
\textit{defend}), 
volleyball (\textit{serve}, \textit{overhand hit}, \textit{underhand hit}, and 
\textit{jump hit}), football (\textit{kick}, 
\textit{block} and \textit{save}), table tennis (\textit{serve}, 
\textit{forehand hit} and \textit{backhand hit}), sprint 
(\textit{run}) and hurdles (\textit{run} and \textit{jump}). 
% Figure~\ref{fig:G3Di} 
% shows example frame of each data modality. 
% In this dataset, volleyball was played collaboratively and the other sports in 
% competitive mode. 
Most sequences contain multiple action classes in a controlled indoor 
environment with a fixed camera. Similar to {G3D}, action point and action segment are provided as ground truth.
%% \begin{figure}[htbp!]
%%    \centering
%% %    \includegraphics[keepaspectratio=true, scale =0.5]{./images/G3Di.png}
%%    \caption{Sample frames of G3Di dataset. From~\cite{Bloom2014}}
%%    \label{fig:G3Di}
%% \end{figure}
\subsubsection{Office Activity dataset}
Office Activity dataset~\cite{Wang2014a}(\url{http://vision.sysu.edu.cn/projects/3d-activity/}) was collected by Sun 
Yat-Sen University in 2014 aimed at complex activities that may typify an 
office environment. Three RGB-D cameras were set up in two scenes and at 
different viewpoints within the scene to capture activities in multiple views. 
The dataset consists of two parts: OA1 
and OA2. In OA1, each activity was performed by a single subject. Five subjects 
performed 10 classes of activities, namely \textit{answering-phones, 
arranging-files, eating, moving-objects, going-to-work, finding-objects, 
mopping, sleeping, taking-water, wandering}. The activities in OA2 are 
interactive activities performed by two subjects, and include \textit{ 
asking-and-away, called-away, carrying, chatting, delivering, 
eating-and-chating, having-guest, seeking-help, shaking-hands, showing}. In 
total, there are 1180 RGB-D activity sequences in Office Activity dataset.
%(see examples in Figure~\ref{fig:Office}).

%\begin{figure}[htbp!]
%   \centering
%   \includegraphics[keepaspectratio=true, scale =0.3]{./images/OfficeActivity.png}   
%   \caption{Sample frames of the Office Activity dataset~\cite{Wang2014a}}
%   \label{fig:Office}
%\end{figure}

\subsubsection{$M^2$I dataset}
The $M^2$I dataset~\cite{Xu2015}(\url{http://media.tju.edu.cn/tju_dataset.html}) was captured by Tianjin 
University in 2015. This dataset contains both human-object interactive actions 
and human-human interactive actions captured from two different views. The 
human-object interactive actions include: \textit{throwing basketball, bouncing 
basketball, twirling hula-hoop, tennis swing, tennis serve, calling cellphone, 
drinking water, taking photos, sweeping the floor, cleaning the desk, playing 
guitar, playing football, passing basketball,} and \textit{carrying box}, where 
the last three actions were performed by two people. The human-human 
interactive actions include: \textit{walking, crossing, waiting, chatting, 
hugging, handshaking, high-fives, bowing,} and \textit{boxing}. Each 
human-object interaction was performed by 22 persons twice and they represent both daily life 
and sport actions. Each human-human interactive action was performed by 20 
groups (two persons in a group) with 2 repetitions. This dataset contains 
1760 action samples in total. The RGB, depth, human mask, and skeleton data are 
all  available. 
\subsubsection{ShakeFive Dataset}
ShakeFive Dataset~\cite{Gemeren2014}(\url{http://www.projects.science.uu.nl/shakefive/}), collected by Universiteit Utrecht in 2014, is a 
dyadic interactions dataset, which contains only two actions, namely 
\textit{hand shake} 
and \textit{high five}. This dataset is aimed at algorithms designed to 
recognize fine-grained interactions and consists of 100 RGB videos 
along with Kinect~\texttrademark skeleton measurements for each 
subject. Fifty-seven videos contain \textit{hand shake} interactions and 43 
contain \textit{high five} 
interactions. 
%Figure~\ref{fig:ShakeFive} shows examples hand shake (left) and 
%high five (right) actions. 
Metafiles provided store the ground truth, which contain 
frame numbers, twenty skeleton joint positions per person, and one of 5 
possible labels describing the interaction in the frame: \textit{standing}, 
\textit{approaching}, \textit{hand shake}, \textit{high five} and 
\textit{leaving}.
% 
%% \begin{figure}[!ht]
%%    \centering
%% %    \includegraphics[keepaspectratio=true, scale =0.5]{./images/ShakeFive.png}
%%    \caption{Sample frames of ShakeFive dataset. From~\cite{Gemeren2014}}
%%    \label{fig:ShakeFive}
%% \end{figure}

\begin{scriptsize}
\begin{longtable}[HTBP]{|p{2.5cm}|p{1.5cm}|P{1.3cm}|P{2.5cm}|P{6.7cm}|}
\hline
Dataset & Year(Cited\footnote{Citations are as of 31 August 2015}) & Modality & \#a,\#s,\#e & Protocol\\
\hline\hline
SBU ~\cite{Yun2012} & 2012(33) & C,D,S & 8,7,300 & 5-fold CV\\
\hline
K3HI ~\cite{Hu2013} & 2013(5) & S & 8,15,320 & 4-fold CV \\
\hline
LIRIS ~\cite{Wolf2014} & 2014(2) & C,D,G & 10,21,828 & 
\setlist{nolistsep,leftmargin=*} \begin{enumerate}[topsep=0pt, partopsep=0pt, 
itemsep=0pt,parsep=0pt]
      \item D1 (305 action samples training, 156 samples test)
      \item D2 (242 action samples training, 125 samples test) \vspace{-1em}
\end{enumerate} \\
\hline
G3Di ~\cite{Bloom2014} & 2014(0) & C,D,S & 6,12,72 & LOSubO \\
\hline
Office Activity ~\cite{Wang2014a} & 2014(0) & C,D,S & 20(10OA1+10OA2), 5,1180 & 5-fold cross validation \\
\hline
$M^2$I TJU ~\cite{Xu2015}& 2015(0) & C,D,S,HM & 22,20,1760 & -\\
\hline
ShakeFive ~\cite{Gemeren2014} & 2014(0) & C,S & 2,37,100 & 
\setlist{nolistsep,leftmargin=*} \begin{enumerate}[topsep=0pt, partopsep=0pt, 
itemsep=0pt,parsep=0pt]
      \item 75\% training (4-fold CV)
      \item 25\% training (4-fold CV) \vspace{-1em}
\end{enumerate} \\
\hline
\captionsetup{width=\textwidth}
\caption{Summary of the key specifications of the human-human 
interaction and multi-person action/activity datasets. Notation for the header: 
\#a: number of actions, \#s: number of subjects, \#e: number of total examples. 
Notation for data format: C: Colour, D: Depth, S: Skeleton, G: Grayscale, HM: 
Human Mask. Notation for protocol: LOSubO: Leave One Subject Out, CV: Cross 
Validation}
\label{table:Basicmultiperson}
\end{longtable}
\end{scriptsize}

\section{Analysis}
\label{sec:analysis}

% This section provides the analysis on the reviewed datasets 
% from several perspectives and provides some recommendations on the future 
% selection of datasets and evaluation protocols.
% 
% To motivate our recommendations, we compare and analyse the
% attributes and characteristics of these datasets, including the potential 
% application scenarios and algorithms that can be tested, complexity level, 
% state-of-the-art results obtained to date, and commonly used evaluation 
% protocols on the reviewed RGB-D action datasets. Based on these analysis, we 
% provide sets of recommendations on both the selection of existing datasets and 
% evaluation protocols.
%The detailed descriptions of the datasets provided in Section~\ref{sec:datasets} form the basis of our analysis. Our discussion 
The analysis presented in this section is framed by consideration for (i) the category of application scenarios, (ii) 
characteristics of dataset acquisition and presentation format, (iii) 
dependence of algorithm evaluation on dataset acquisition modes, (iv) complexity 
of the environmental factors inherent in dataset, (v) evaluation protocols 
commonly used for algorithm development and testing,  and (vi) state-of-the-art 
results obtained to date with the datasets. Naturally, the discussions invite 
some recommendations and they are provided appropriately.
\subsection{Application scenarios}
\label{sec:application}
The creation of a given dataset is usually motivated and targeted at 
some real-world applications. Lun et al.~\cite{Lun2015} summarized the major applications from the algorithm development perspective in~\cite{Lun2015}. In this paper, two broad categories of applications are identified and 
they are characterized by the types of actions in the dataset or the description 
provided by the dataset creators. The first category is human-computer 
interaction (HCI), example applications include video game interface and device 
control. The second category is daily activity (DA), including scene surveillance, 
elderly monitoring, service robotics, E-healthcare and smart rooms. 
%elderly monitoring, sport analysis, telehealth, biometrics and smart rooms. 
Ostensibly, the various datasets model the applications well, but the various 
environmental factors and the size of examples need to be considered in 
determining how well a dataset mimics reality.
Table~\ref{table:application} (columns one and two) presents a summary of the 
datasets reviewed and the target applications.  

% The applications of a dataset are generally consider in two 
% aspects: the real world application and algorithm evaluation. 
% Table~\ref{table:application} provides a list of real world applications and 
% algorithms that can be evaluated using the reviewed datasets. The list is given 
% in chronological order of the release of datasets. Below, we detailed analyse 
% the applications in terms of both real world and algorithm evaluation.
% 
% \subsubsection{Real world applications}
% All the datasets are designed to mimic particular real world scenarios. 
% Hence, we summarise the categories of real world applications of each dataset. 
% The categorises of real world applications are classified depending on the type 
% of involved action classes in a dataset or the descriptions provided by the 
% original authors who released the dataset. Two main categorizes can be 
% identified: 
% Human-Computer Interaction (HCI), including video game, device control, 
% etc.; Daily Activity (DA), including surveillance, cooking, elderly monitoring, 
% sports analysis, telehealth, biometrics, smart rooms, etc.
% The applications of current RGB-D-based action datasets are still very limited because of the limited types of action involved in each dataset. Most of RGB-D datasets are collected in lab environment and the execution of actions is generally follow the instructors instruction. Some even follow a same performing style. In this case, even different subjects are involved, the variation of performing style are subtle. 
\subsection{Characteristics of dataset acquisition}
\label{sec:characteristics_dataset}
The characteristics of the dataset acquisition modes and the presentation format 
has bearing on how algorithms can use them for evaluation without repurposing. 
A set of \textit{de facto} standard acquisition modes and presentation formats 
potentially provide a basis for objective comparative evaluation of algorithms. 
Based on the datasets reviewed, four modes of acquisition and presentation 
along with two modes that are variations of the third and fourth modes can be identified.  
They are listed below with some explanations:
% \subsubsection{Algorithm Evaluation}
% The research domain of most datasets is action recognition, apart from 
% some that may be applicable to action detection, falling detection, online 
% action recognition and object tracking. The algorithms that a dataset can 
% evaluate depend on the acquisition of data and the provided ground truth label 
% of a dataset. Based on the reviewed datasets, we define and summarise four modes 
% of data acquisition and storages.
\begin{itemize}[topsep=0pt, partopsep=0pt, itemsep=0pt,parsep=0pt]
\item Mode 1: Captured as action samples and stored in segments where
each segment contains only one action or activity.
\item Mode 2: Captured as activity samples, but each activity contains a 
continuous sequence of labelled sub-activities.
\item Mode 3: Captured as sequences of actions where the order of the 
actions in each sequence is fixed. The data is stored in sequential fashion 
and action segment points are provided.
\item Mode 4: Captured as sequences of actions where the order of actions 
in each sequence is random. The data are stored in sequential fashion and 
action segment points are provided.
\item Mode 3*: Captured as in Mode 3, but stored and presented as in 
Mode 1 after some processing.
\item Mode 4*: Captured as in Mode 4, but stored and presented as in 
Mode 1 after some processing.
\end{itemize}
Table~\ref{table:application} (columns one and four) presents a summary of the 
datasets reviewed and the acquisition mode.
%1 86 0 1 0 1 walking
%48 76 1 0 0 0 waving hand
%
%In this case, the first row indicates a "walking" action executed by right and left legs, from frame 1 to 86. The second row, a "waving hand" action executed with right arm, between frames 48 and 76. 
%
%It is worth to note that the maximum frame number of every action stamp is always lower or equal than the minimum number of frames of the skeleton files, and could be lower than the actual number of frames of the video. 

\subsection{Algorithm evaluation and dataset acquisition modes}
The development and implementation of a given application may require 
several algorithms and these will need to be evaluated objectively. Based on 
the acquisition and presentation modes, and available ground truth labels, the 
datasets can be used for testing five identifiable types of algorithms. These 
include action recognition, action detection, falling detection and online 
action recognition. Detailed explanations are provided as 
follows.
\begin{description}[topsep=0pt, partopsep=0pt, itemsep=0pt,parsep=0pt]
\item [Action Recognition:] In this paper, action recognition and action 
categorization are synonymous and we assume that a unique label can 
represent the entire video sequence. This casts the human action 
recognition problem as a classification problem.%~\cite{Poppe2010}. 
Datasets captured and presented in Mode 1, as well as Mode 3* and Mode 4*, can be 
directly used for 
action recognition. The datasets presented in other modes can also be used 
for action recognition after some processing, e.g. segmenting sequence into action 
samples using the ground truth action segment points.
\item [Action Detection:] This
focuses on identifying the occurrence of specific actions in an 
observed sequence.%~\cite{Poppe2010}. 
Thus, to test action detection algorithms 
the dataset should be captured continuously and provide accurate ground truth 
segmentation points of each action. Only the datasets captured in Modes 2, 3 
and 4 can be used for action detection.
\item [Falling Detection:] This is an important but specific type of
action detection which only focuses on falling event. Its importance has 
risen because of the potential application in health monitoring. A dataset 
meant for the evaluation of \textit{falling detection} algorithm should be 
captured in similar modes as \textit{action detection} but should also 
contain falling events and possibly other actions that are easily 
confused with falling actions.
\item [Online Action Recognition:] For the evaluation of online action 
recognition algorithms, the dataset must mimic the realistic scenario where 
unlabeled video sequence are continuously presented. Additionally, 
the actions should also be performed in random order. Datasets captured in Mode 
4 are the only ones suitable for this category of algorithms.
% \item [Object Tracking:] Previous works have shown that how an 
% object is being used is assistive for action recognition 
% tasks~\cite{Gupta2009}~\cite{Yao2010}~\cite{Koppula2013a}. Thus, many algorithms 
% perform object tracking to assist action recognition. To evaluate the object 
% tracking, a dataset should contain human-object interaction and the ground truth 
% locations and labels of the involved objects should be provided.
\end{description}

\begin{table}[!hptb]
\begin{center}
\begin{scriptsize}
\begin{tabular}{|m{2.6cm}<{\centering}|c|m{3.2cm}<{\centering}|m{4.2cm}<{\centering}|m{2.7cm}<{\centering}|}
\hline
\textbf{Single view} & \textbf{Applications} & \textbf{Algorithm Evaluation} & \textbf{Data acquisition/presentation} & \textbf{Ground truth}\\
\hline
MSRAction3D & HCI & AR & Mode 1 & AN\\
RGBD-HuDaAct & DA & AR & Mode 1 & AN\\
CAD-60 & DA & AR/AD & Mode 1 & AN\\
MSRC-12 & HCI & AR & Mode 3 & AN/ASP/TD \\
MSRDaily & DA & AR & Mode 1 & AN\\
UTKinect & HCI & AR & Mode 3 & AN/ASP \\
G3D & HCI & AR & Mode 2 & AN/SAN/SASP \\
DHA & HCI & AR & Mode 1 & AN \\
Falling Event & DA & FD & Mode 1 & AN\\
MSRActionPair & DA & AR & Mode 1 & AN\\
CAD-120 & DA & AR/AD/ObT & Mode 2 & AN/SAN/SASP/OL \\
WorkoutSU-10 & DA & AR & Mode 1 & AN/TD \\
Concurrent Action & DA & AR/OAR & Mode 4 & AN/ASP \\
IAS-lab & DA & AR & Mode 1 & AN\\
UCFKinect & HCI & AR & Mode 1 & AN\\
Osaka & HCI & AR  & Mode 1 & AN\\
Morning-Routine & DA & AR/AD/ObT & Mode 3 & AN/ASP/ASL\\
RGBD-SAR & DA & AR & Mode 1 & AN\\
Mivia & DA & AR & Mode 1 & AN\\
UPCV & DA & AR & Mode 1 & AN\\
TJU & HCI & AR & Mode 1 & AN \\
MAD & HCI & AR/AD & Mode 3 & AN/ASP\\
Composable & DA & AR/AD & Mode 2 & AN/SAN/SASP/ArLg\\
3D Online & DA & AR/OAR/ObT & Mode 1/ Mode 4 & AN/OL/ASP \\
RGB-D activity & DA & AR/AD & Mode 4 & AN per frame \\
UTD-MHAD & HCI & AR & Mode 1 & AN \\
SYSU & DA & AR & Mode 1 & - \\
\hline \hline
\textbf{Multi-view} & \textbf{Applications} & \textbf{Algorithm Evaluation} & \textbf{Data acquisition/presentation} & \textbf{Ground truth}\\
\hline
ATC$4^2$ & DA & AR/FD & Mode 1 & AN\\
Falling Detection & DA & FD & Mode 4 & FSP\\
Berkeley MHAD & HCI & AR & Mode 1 & AN\\
DMLSmartActions & DA & AR/OAR & Mode 4 & AN/ASP\\
ReadingAct & DA & AR & Mode 1 & -\\
Multiview 3D Event & DA & AR/AD/ObT & Mode 3* & AN/OL\\
Northwestern-UCLA & DA & AR & Mode 1 & AN\\
UWA3D Multiview & DA/HCI & AR & Mode 4* & AN\\
Multi-view TJU & HCI & AR  & Mode 1 & AN\\
NJUST & HCI & AR & Mode 1 & AN \\
\hline \hline
\textbf{Multi-person} & \textbf{Applications} & \textbf{Algorithm Evaluation} & \textbf{Data acquisition/presentation} & \textbf{Ground truth}\\
\hline
SBU  & DA & AR & Mode 1 & AN\\
K3HI & DA & AR & Mode 1 & AN\\
LIRIS & DA & AR & Mode 1 & AN/ASL\\
G3Di & HCI & AR/AD  & Mode 3 & AN/ASP/AP\\
Office Activity & DA & AR & Mode 1 & AN\\
3M TJU & DA/HCI & AR & Mode 1 & AN \\
ShakeFive & DA & AR & Mode 1 & AN\\

\hline
\end{tabular}
\end{scriptsize}
\end{center}
\vspace{-2em}
\caption{Real world applications and algorithm evaluations. 
Notation for real world application: DA: Daily Activity; HCI: Human Computer 
Interaction. Notation for algorithm evaluations: AR: Action Recognition; ObT: 
Object Tracking; AD: Action Detection; {OAR}: Online Action Recognition; FD: 
Falling Detection. Notation for ground truth: AN: Action Name; ASP: Action 
Segment Point; TD: Text Description; SAN: Sub Action Label; SASP: Sub Action 
Segment Point; FSP: Falling Segment Point; ASL: Actor Spatial Location; ArLg: 
right or left Arm, right or left Leg; OL: Object Location; AP: Action Point. }
\label{table:application}
\end{table}

\subsection{Complexity of the environmental factors inherent in datasets}
\label{sec:complexity}
% \subsubsection{Challenging factors of datasets}
The comparative performance of a given algorithm depends on the 
environmental factors that are represented in the dataset being used for evaluation. 
Incidentally, the degree of complexity of the factors should also be 
considered. For example, a dataset with fixed but cluttered background may not 
be as challenging as one where the cluttered background varies from sample to 
sample. To judge the degree of challenege posed by a dataset consideration 
should be given to the complexity of the actions performed and the attending 
environmental factors. Ramanathan et al.~\cite{Ramanathan2014} identified 
some of these factors as execution rate, anthropomorphic variations, viewpoint 
variation, occlusion, cluttered background, and camera motion. In order to 
evaluate an algorithm targeted at real-world applications, a good dataset should 
represent some of these factors and exercise the robustness of the algorithm. 
Ideally, the dataset should model the real-world application.

Most of the reviewed RGB-D datasets include execution rate and anthropomorphic 
variations to some extent, since these factors can be achieved by employing different 
individuals and several repetition. However, viewpoint variation is only found 
in multi-view dataset. Only small subset of the datasets include occlusion and 
cluttered background. The lack of occlusion and acquisition in relatively 
simple background limits the usefulness of any dataset in the design 
of realistic algorithms. Camera motion is not frequently found in
RGB-D-based action datasets. Although the location and orientation of camera 
were not fixed in DMLSmartActions dataset, the camera was static during data 
capture and cannot be regarded as camera motion. Only LIRIS dataset incorporates 
camera motion because the camera was mounted on a mobile robot. Apart from these 
common challenges that are also typical of 2D video datasets, another issue 
related to RGB-D-based action dataset is the useful range (for depth data) of 
the Kinect~\texttrademark camera. This limitation has restricted the capture 
environment to indoors and hence also limits the usefulness of these datasets 
in testing algorithms meant to operate outdoor.

% In this paper, we take six key challenging factors that 
% usually affect the recognition difficulty into consideration: background clutter 
% and occlusion, kinematic complexity of the actions/activities, variability 
% amongst the actions/activities within a dataset, variations of the action being 
% performed execution speed and personal style, \textit{compositional actions}, 
% and interaction between human and objects. For the compositional actions, we 
% define a compositional action as one action composed of two or more actions, 
% which are recognisable actions in their own right. For example, \textit{pick 
% up}\& \textit{throw} and \textit{high throw} are two individual actions 
% contained in MSR Action 3D dataset, but \textit{pick up}\& \textit{throw} 
% contains \textit{high throw}, which makes them confusable actions. With or 
% without human-object interaction is another important characteristic of a 
% dataset, because as mentioned in Section~\ref{sec:application}, some algorithms 
% may benefit from the objects that the actors interact with. 

% \subsubsection{Complexity level}
It is instructive to describe and assign level of complexity to a selection 
of these factors: background clutter and occlusion, kinematic complexity 
of the actions/activities, variability amongst the actions/activities within a 
dataset, execution speed and personal 
style, \textit{composable actions}, and interactivity between human and 
objects. We define a composable action as one 
composed of two or more actions, which are recognisable actions in their own 
right. For example, \textit{pick 
up}\& \textit{throw} and \textit{high throw} are two individual actions 
contained in MSR Action 3D dataset, but \textit{pick up}\& \textit{throw} 
contains \textit{high throw}, which makes them confusable actions. Human-object 
interactivity is another important characteristic of a 
dataset because some algorithms may benefit from the objects that the actors 
interact with~\cite{Gupta2009,Yao2010,Koppula2013a}. 

Table~\ref{table:complexity} summarizes the assignment of the level of 
complexity  of environmental factors found in the datasets reviewed. 
The order of datasets are in chronological order. The first four factors 
could take on one of three levels of complexity (low, medium, and 
high) while the last two are binary valued (yes/no). The 
criteria for categorization are summarized as follows.

\noindent
% \begin{itemize}
\textbf{Background clutter and occlusion}
\begin{itemize}[topsep=0pt, partopsep=0pt, itemsep=0pt,parsep=0pt]
\item Low: the background is fixed and clean. There is no occlusion of the 
subjects.
\item Medium: the background is fixed but is cluttered. Some occlusion of 
subjects may be present.
\item High: the background is not fixed among action samples and/or is 
cluttered. Occlusions are present and the actions may be affected by the 
background and occlusion.
\end{itemize}
\textbf{Kinematic complexity}
\begin{itemize}[topsep=0pt, partopsep=0pt, itemsep=0pt,parsep=0pt]
\item Low: the movements are relatively simple and with short duration.
\item Medium: the movements are of medium complexity and the duration is 
longer than movements in the low level category.
\item High: the movements are complex and with long duration.
\end{itemize}
\textbf{Variability amongst actions}
\begin{itemize}[topsep=0pt, partopsep=0pt, itemsep=0pt,parsep=0pt]
\item Low: the variation of complexity levels amongst actions within a dataset is low.
\item Medium: the variation of complexity levels amongst actions within a dataset is medium.
\item High: the variation of complexity levels amongst actions within a dataset is high.
\end{itemize}
\textbf{Execution rate}
\begin{itemize}[topsep=0pt, partopsep=0pt, itemsep=0pt,parsep=0pt]
\item Low: the variation in style of execution among different subjects 
or repetitions is low
\item Medium: the variation in style of execution among different subjects or 
repetitions is medium.
\item High: the variation in style of execution among different subjects or 
repetitions is high.
\end{itemize}
\textbf{Composable actions}: whether a dataset contain composable 
actions (Yes/No).\\
\textbf{Human-object interaction}: whether a dataset contain human-object 
interaction (Yes/No).
% \end{itemize}
% %For the background and occlusion factor, if the background is 
% fixed with little clutter and the action is not occluded by other objects, the 
% complexity level of this factor is low. In contrast, if the action is performed 
% in a cluttered environment or with human-object occlusion, the complexity level 
% is considered to be high. The kinematic complexity refers to the overall 
% complexity level of actions involved in a dataset. The variability amongst 
% actions means the difference degree of complexity levels among actions within a 
% dataset, which can also significantly affect the difficulty of action 
% recognition task. 

\begin{table}[!hpt]
\scriptsize
\begin{center}
\begin{scriptsize}
\begin{tabular}{|m{2.8cm}<{\centering}|m{2cm}<{\centering}|m{1.8cm}<{\centering}|m{2.4cm}<{\centering}|m{1.3cm}<{\centering}|m{1.8cm}<{\centering}|c|}
\hline
\textbf{Single view} & \textbf{Background\& occlusion} & \textbf{Kinematic
complexity} & \textbf{Variability amongst actions} & \textbf{Execution rate} & 
\textbf{Composable actions} & \textbf{Object}\\
\hline\hline
MSRAction3D & Low & Low & Low & Low & Yes & No \\
%pickup & through contains high throw
\hline
RGBD-HuDaAct & Medium & Medium & Medium & High & Yes & Yes\\
%eat meal and make a phone call are relatively complex
\hline
CAD-60 & Medium & Medium & Medium & Low & Yes & Yes\\
%rinsing mouth contain rinsing mouth
\hline
MSRC-12& No background & Low & Low & Medium & No & No \\
\hline
MSRDaily & Medium & High & Low & Low & No & Yes \\
%many actions are small and similar
\hline
UTKinect & Medium & Medium & Low & Medium & Yes & Yes\\
%carry contains walking
\hline
G3D & Low & Low & Low & Low & No  & No \\
\hline
DHA & Low & Low & Low & Low & No  & No \\
\hline
Falling Event & Low & Low & Low & Low & Yes & No \\
\hline
MSRActionPair & Low & Low & Low & Low & No & Yes \\
\hline
CAD-120 & High & High & High & Medium & No & Yes\\
%picking objects, taking food are relatively easy, making cereal and stacking 
%objects etc are relatively complex
\hline
WorkoutSU-12 & Low & Low & Low & Low & No & No \\
\hline
Concurrent Action & No background & Medium & High & High & No & No \\
\hline
IAS-lab & Low & Low & Low & Low & Yes & Yes\\
\hline
UCFKinect & No background & Low & Low & Low & No & No \\
\hline
Osaka & Low & Low & Low & Low & No & No \\
\hline
Morning-Routine & Medium & High & Medium & Only one subject & No & Yes \\
\hline
RGBD-SAR & High & High & Medium & High & No & Yes \\
\hline
Mivia & Low & Low & Low & Low & No & Yes \\
\hline
UPCV & No background & Low & Low & Low & No & No \\
\hline
TJU & Low & Low & Low & Low & No & No \\
\hline
MAD & Low & Low & Low & Low  & No & No \\
\hline
Composable & Low & High & Medium & High & Yes & Yes \\
\hline
3D Online & Medium & Medium & Low & Medium & No & Yes \\
\hline
RGB-D activity & High & High & High & High & Yes & Yes \\
\hline
UTD-MHAD & Low & Low & Low & Low & Yes & No \\
\hline
SYSU & Not released & - & - & - & - & - \\
\hline

\textbf{Multi view} & \textbf{Background\& occlusion} & \textbf{Kinematic complexity} & \textbf{Variability amongst actions} & \textbf{Execution rate} & \textbf{Compositional actions} & \textbf{Object}\\
\hline\hline
ATC$4^2$ & Low & Low & Low & Low & No & Yes \\
\hline
Falling Detection & High & Low & Low & Medium & Yes & Yes \\
\hline
Berkeley MHAD & Low & Low & Low & Low & No & No \\
\hline
DMLSmart & High & Low & Low & Medium & Yes & Yes \\
\hline
ReadingAct & Not released & - & - & - & - & - \\
\hline
Multiview 3D Event & High & Medium & Low & Low & No & Yes \\
\hline
Northwestern-UCLA & Low & Low & Low & Low & No & Yes \\
\hline
UWA3D Multiview & Low & Low & Low & Low & Yes & No \\
\hline
Multi-view TJU & Low & Low & Low & Low & No & No \\
\hline
NJUST & Low & Low & Low & Low & No & No \\
\hline
\textbf{Multi person} & \textbf{Background\& occlusion} & \textbf{Kinematic complexity} & \textbf{Variability amongst actions} & \textbf{Execution rate} & \textbf{Compositional actions} & \textbf{Object}\\
\hline\hline
SBU  & Low & Low & Low & Low & No & No \\
\hline
LIRIS & High & High & High & High & Yes & Yes \\
\hline
K3HI & No background & Low & Low & Low & No & No \\
\hline
G3Di & Low & High & Low & Medium & No & No \\
\hline
Office Activity & High & Medium & Medium & High & No & Yes \\
\hline
$M^2$I TJU & Medium & Medium & Low & Low & No & Yes \\
\hline
ShakeFive & Low & Low & Low & Low & No & No \\
\hline

\end{tabular}
\end{scriptsize}
\end{center}
\vspace{-2.8em}
\caption{Complexity level of the reviewed datasets from different aspects }
\label{table:complexity}
\end{table}
\subsection{Evaluation protocols}
\label{sec:evaluationProtocol}
%
%4.4 and 4.5 should be rewritten by following the logic of ICCV paper. 
%
Careful design of the evaluation protocols is necessary to validate the results 
reported for each algorithm. Also important is the matching of the algorithm 
insofar as its purpose can be articulated, with the dataset represnting the 
enviromental factors that underpin the purpose. 
Several algorithms have been evaluated using the datasets reviewed 
in this paper. Using the algorithms that reported state-of-the-art results as a 
basis, a number of evaluation setup are found to be in common usage. They are 
listed and described below:
% Based on the reviewed algorithms with state-of-the-art results, the most 
% popular set-ups can be identified as follows:
% The results obtained from the algorithms that were evaluated on the 
% reviewed datasets depend on the algorithm and more importantly on the 
% protocols used for training and test. It is important to match the algorithm, 
% dataset and the evaluation 
% protocol in order to report valid results. 
% Based on the reviewed algorithms with state-of-the-art results, the most 
% popular
\begin{description}[topsep=0pt, partopsep=0pt, itemsep=0pt,parsep=0pt]
\item [Leave-one-sequence-out cross validation setup:]
Randomly select one sequence from the entire dataset as test data and use the 
remaining sequences as training data. Perform a certain number of these 
tests and average the outcomes as the final result.
\item [Leave-one-subject-out cross validation setup:]
Train with all but one subject and test with the unseen data. Repeat this 
for all subjects and report the average of the outcomes as the final result.
\item [Cross-subject test:]
A number of the subjects are used for training and the remainder for testing.
\begin{itemize}[topsep=0pt, partopsep=0pt, itemsep=0pt,parsep=0pt]
\item Select half of the subjects to be used for training and the remainder
for testing. Some may use two-fold cross validation: repeat the evaluation 
using the previous test set as the training set and vice versa. The final 
result is the average of the two tests.
\item Consider all the possible combinations of half subjects for training and 
the remaining for test.
\end{itemize}
%\item One-shot learning\\
%Only one sequence in each action needs to be used for training and other 
%sequences are used for testing. 
\item [Cross-view:]
Select one view as training set and the other views as test set. This only 
applies to multi-view datasets. 
\item [Cross-environment:]
Select the actions performed in one environment as 
training and test on actions performed in other environments. 
This is only applicable to datasets with specific actions captured in 
different environments. 
\end{description}

\subsection{State-of-the-art results}
\label{sec:State-of-the-art}
In this section, we tabulate the state-of-the-art methods\footnote{Authors will maintain a website to keep updating the state-of-the-art results} that used the reviewed 
datasets in order to highlight current status of research. For most of the 
datasets, we provide more than one algorithm because, not having used 
the same evaluation protocol, the qualifier ``state-of-the-art" is not 
unequivocal. In addition, even when the same datasets and evaluation protocols 
have been used, the data modalities also need to be taken into consideration. 
This important observation has previously been ignored by researchers. 
There are instances where algorithms have been tested on skeleton 
data and claim of superior performance made over algorithms tested on depth 
data. In Tables~\ref{table:ResultsSingle}, 
\ref{table:ResultsMultiview} and \ref{table:ResultsMultiperson}, we provide the 
state-of-the-art methods along with the reported results, the modalities of the 
algorithm used, and the protocol used for training and evaluation of the 
algorithms. The listing is in descending order of citation 
frequency of the original paper that published the datasets. 

%\newpage
\begin{scriptsize}
\begin{longtable}[HTBP]{|p{2cm}|P{3.8cm}|P{4cm}|P{1.1cm}|P{4.2cm}|}

\hline
Dataset & State-of-the-art Methods & Acc.(\%) & Data & Protocol\\
\hline\hline

MSR-Action3D ~\cite{Li2010} & 
\setlist{nolistsep,leftmargin=*} \setlist{nolistsep,leftmargin=*} 
\begin{enumerate}[topsep=0pt, partopsep=0pt, itemsep=0pt,parsep=0pt]
      \item TriViews +PFA ~\cite{Chen2015}
      \item Decision-Level Fusion (SUM)~\cite{Zhu2015}
      \item ConvNets ~\cite{pichao2015,pichaoTHMS} \vspace{-1em}
\end{enumerate} & \setlist{nolistsep,leftmargin=*} \begin{enumerate}[topsep=0pt, 
partopsep=0pt, itemsep=0pt,parsep=0pt]
      \item 98.2
      \item 98.2 
      \item 100 \vspace{-1em}
\end{enumerate}
 & \setlist{nolistsep,leftmargin=*} \begin{enumerate}[topsep=0pt, partopsep=0pt, 
itemsep=0pt,parsep=0pt]
      \item D, S
      \item D, S
      \item D  \vspace{-1em}
\end{enumerate} 
& \setlist{nolistsep,leftmargin=*} \begin{enumerate}[topsep=0pt, partopsep=0pt, 
itemsep=0pt,parsep=0pt]
      \item CS (Half training, half test) 
      \item CS (2,3,5,7,9 subject training, 1,4,6,8,10 subject test)
      \item CS (1,3,5,7,9 subject training, 2,4,6,8,10 subject test) \vspace{-1em}
\end{enumerate} \\
\hline
MSRDaily-Activity3D ~\cite{Wang2012} & 
\setlist{nolistsep,leftmargin=*} \begin{enumerate}[topsep=0pt, partopsep=0pt, 
itemsep=0pt,parsep=0pt]
      \item $\tau$-test~\cite{Lu2014}
      \item DL-GSGC +TPM ~\cite{Luo2013}
      \item 3D joint+CS-MLtp~\cite{Luo2014}
      \item Depth-VSFR~\cite{Cho2015} \vspace{-1em}
\end{enumerate} & \setlist{nolistsep,leftmargin=*} \begin{enumerate}[topsep=0pt, 
partopsep=0pt, itemsep=0pt,parsep=0pt]
      \item 95.63
      \item 95
      \item 92.5
      \item 89.7 \vspace{-1em}
\end{enumerate}
 & \setlist{nolistsep,leftmargin=*} \begin{enumerate}[topsep=0pt, partopsep=0pt, 
itemsep=0pt,parsep=0pt]
      \item D,S
      \item S
      \item C,S
      \item D \vspace{-1em}
\end{enumerate} 
& \setlist{nolistsep,leftmargin=*} \begin{enumerate}[topsep=0pt, partopsep=0pt, 
itemsep=0pt,parsep=0pt]
      \item Not given
      \item CS (Half training, half test) 
      \item CS (Half training, half test)
      \item Not given \vspace{-1em} 
\end{enumerate} \\
\hline
UTKinect ~\cite{Xia2012}  & 
\setlist{nolistsep,leftmargin=*} \begin{enumerate}[topsep=0pt, partopsep=0pt, 
itemsep=0pt,parsep=0pt]
      \item Fused feature~\cite{Ye2015}    
      \item TriViews +PFA ~\cite{Chen2015}
      \item Grassman manifold~\cite{Slama2014}  \vspace{-1em} 
\end{enumerate} & \setlist{nolistsep,leftmargin=*} \begin{enumerate}[topsep=0pt, 
partopsep=0pt, itemsep=0pt,parsep=0pt]
	  \item 100
      \item 98
      \item 95.25 \vspace{-1em}    
\end{enumerate}
 & \setlist{nolistsep,leftmargin=*} \begin{enumerate}[topsep=0pt, partopsep=0pt, 
itemsep=0pt,parsep=0pt]
      \item C, D, S      
      \item D, S
      \item D \vspace{-1em}
\end{enumerate} 
& \setlist{nolistsep,leftmargin=*} \begin{enumerate}[topsep=0pt, partopsep=0pt, 
itemsep=0pt,parsep=0pt]    
      \item CS (Half training, half test)
      \item CS (Half training, half test)
      \item LOSubO \vspace{-1em}
\end{enumerate} \\
\hline
CAD-60 ~\cite{Sung2011} & 
\setlist{nolistsep,leftmargin=*} \begin{enumerate}[topsep=0pt, partopsep=0pt, 
itemsep=0pt,parsep=0pt]
      \item Decision-Level Fusion (Majority Voting) ~\cite{Zhu2015}      
      \item Pose Kinectic Energy~\cite{Shan2014}
      \item SpatioTemporal Interest Pt~\cite{Zhu2014} \vspace{-1em}
\end{enumerate} & \setlist{nolistsep,leftmargin=*} \begin{enumerate}[topsep=0pt, 
partopsep=0pt, itemsep=0pt,parsep=0pt]
      \item 96.4(Prec.) 84.6(Rec.)
      \item 93.8(Prec.) 94.5(Rec.)
      \item 93.2(Prec.) 84.6(Rec.) \vspace{-1em}
\end{enumerate}
 & \setlist{nolistsep,leftmargin=*} \begin{enumerate}[topsep=0pt, partopsep=0pt, 
itemsep=0pt,parsep=0pt]
      \item D, S
      \item S
      \item D \vspace{-1em}
\end{enumerate} 
& \setlist{nolistsep,leftmargin=*} \begin{enumerate}[topsep=0pt, partopsep=0pt, 
itemsep=0pt,parsep=0pt]
      \item LOSubO(1,3,4 training, 2 test)
      \item LOSubO
      \item LOSubO \vspace{-1em}
\end{enumerate} \\
\hline
RGBD-HuDaAct ~\cite{Ni2011} &
\setlist{nolistsep,leftmargin=*} \begin{enumerate}[topsep=0pt, partopsep=0pt, 
itemsep=0pt,parsep=0pt]
      \item BoW-Pyramid~\cite{Tsai2013}
      \item PA-Pooling~\cite{Ni2014} \vspace{-1em}
\end{enumerate} & \setlist{nolistsep,leftmargin=*} \begin{enumerate}[topsep=0pt, 
partopsep=0pt, itemsep=0pt,parsep=0pt]
      \item 91.7
      \item 85.9 \vspace{-1em}
\end{enumerate}
 & \setlist{nolistsep,leftmargin=*} \begin{enumerate}[topsep=0pt, partopsep=0pt, 
itemsep=0pt,parsep=0pt]
      \item C,D
      \item C \vspace{-1em}
\end{enumerate} 
& \setlist{nolistsep,leftmargin=*} \begin{enumerate}[topsep=0pt, partopsep=0pt, 
itemsep=0pt,parsep=0pt]
      \item LOSubO
      \item LOSubO \vspace{-1em}
\end{enumerate} \\
\hline
MSRAction-Pair ~\cite{Oreifej2013}  & 
\setlist{nolistsep,leftmargin=*} \begin{enumerate}[topsep=0pt, partopsep=0pt, 
itemsep=0pt,parsep=0pt]
      \item BHIM~\cite{Kong2015}
      \item 3D Pose~\cite{Eweiwi2015}
      \item SNV~\cite{Yang2014} \vspace{-1em}
\end{enumerate} & \setlist{nolistsep,leftmargin=*} \begin{enumerate}[topsep=0pt, 
partopsep=0pt, itemsep=0pt,parsep=0pt]
      \item 100
      \item 99.4
      \item 98.89 \vspace{-1em}
\end{enumerate}
 & \setlist{nolistsep,leftmargin=*} \begin{enumerate}[topsep=0pt, partopsep=0pt, 
itemsep=0pt,parsep=0pt]
      \item C, D
      \item S
      \item D \vspace{-1em}
\end{enumerate} 
& \setlist{nolistsep,leftmargin=*} \begin{enumerate}[topsep=0pt, partopsep=0pt, 
itemsep=0pt,parsep=0pt]
      \item CS (First 5 test, rest training)
      \item CS (Odd subjects training, even subjects test)
      \item CS (First 5 test, rest training) \vspace{-1em}
\end{enumerate} \\
\hline

MSRC-12 gesture ~\cite{Fothergill2012}   & 
\setlist{nolistsep,leftmargin=*} \begin{enumerate}[topsep=0pt, partopsep=0pt, 
itemsep=0pt,parsep=0pt]
      \item RDF-selected features~\cite{Negin2013}
      \item Cov3DJ~\cite{Hussein2013}
      \item ESM(6 iconic gestures)~\cite{Jung2015} \vspace{-1em}
\end{enumerate} & \setlist{nolistsep,leftmargin=*} \begin{enumerate}[topsep=0pt, 
partopsep=0pt, itemsep=0pt,parsep=0pt]
      \item 94.03
      \item 93.6 \& 91.7
      \item 96.76 \vspace{-1em}
\end{enumerate}
 & \setlist{nolistsep,leftmargin=*} \begin{enumerate}[topsep=0pt, partopsep=0pt, 
itemsep=0pt,parsep=0pt]
      \item S
      \item S
      \item S \vspace{-1em}
\end{enumerate} 
& \setlist{nolistsep,leftmargin=*} \begin{enumerate}[topsep=0pt, partopsep=0pt, 
itemsep=0pt,parsep=0pt]
      \item LOSubO(5-fold CV)
      \item LOSubO(30-fold CV) \&CS (half subjects training)
      \item LOSubO \vspace{-1em}
\end{enumerate} \\
\hline

CAD-120 ~\cite{Koppula2013a} & 
\setlist{nolistsep,leftmargin=*} \begin{enumerate}[topsep=0pt, partopsep=0pt, 
itemsep=0pt,parsep=0pt]
      \item QQSTR-gt-tracks~\cite{Tayyub2015}
      \item Skeleton feature+HMMs~\cite{Taha2015}
      \item ATCRF~\cite{Koppula2013} \vspace{-1em}
\end{enumerate} & \setlist{nolistsep,leftmargin=*} \begin{enumerate}[topsep=0pt, 
partopsep=0pt, itemsep=0pt,parsep=0pt]
      \item 95.2(Activity Acc.) 95.2(Activity Prec.)95(Activity Rec.)
      \item 94.4(Activity Acc.) 91.6(Sub-activity Acc.)
      \item 93.5(Activity Acc.) 95(Activity Prec.) 93.3(Activity Rec.) 89.3(Sub-activity Acc.) \vspace{-1em}
\end{enumerate}
 & \setlist{nolistsep,leftmargin=*} \begin{enumerate}[topsep=0pt, partopsep=0pt, 
itemsep=0pt,parsep=0pt]
      \item S
      \item S
      \item S \vspace{-1em}
\end{enumerate} 
& \setlist{nolistsep,leftmargin=*} \begin{enumerate}[topsep=0pt, partopsep=0pt, 
itemsep=0pt,parsep=0pt]
      \item LOSubO (4-fold CV)
      \item LOSubO (4-fold CV)
      \item LOSubO (4-fold CV) \vspace{-1em}
\end{enumerate} \\
\hline
UCFKinect ~\cite{Ellis2013} & 
\setlist{nolistsep,leftmargin=*} \begin{enumerate}[topsep=0pt, partopsep=0pt, 
itemsep=0pt,parsep=0pt]
      \item MvMF-HMM~\cite{Beh2014}      
      \item Hierarchical model~\cite{Jiang2013}
      \item Moving Pose~\cite{Zanfir2013} \vspace{-1em}
\end{enumerate} & \setlist{nolistsep,leftmargin=*} \begin{enumerate}[topsep=0pt, 
partopsep=0pt, itemsep=0pt,parsep=0pt]
      \item 98.9
      \item 98.7
      \item 98.5 \vspace{-1em}
\end{enumerate}
 & \setlist{nolistsep,leftmargin=*} \begin{enumerate}[topsep=0pt, partopsep=0pt, 
itemsep=0pt,parsep=0pt]
      \item S
      \item S
      \item S \vspace{-1em}
\end{enumerate} 
& \setlist{nolistsep,leftmargin=*} \begin{enumerate}[topsep=0pt, partopsep=0pt, 
itemsep=0pt,parsep=0pt]
      \item 4-fold CV
      \item 2-fold CV 
      \item 4-fold CV \vspace{-1em}
\end{enumerate} \\
\hline

G3D ~\cite{Bloom2012, Bloom2013} & 
\setlist{nolistsep,leftmargin=*} \begin{enumerate}[topsep=0pt, partopsep=0pt, 
itemsep=0pt,parsep=0pt]
      \item LRBM~\cite{Nie2015}  
      \item Clustered Action Manifolds~\cite{Bloom2014a} \vspace{-1em}
\end{enumerate} & \setlist{nolistsep,leftmargin=*} \begin{enumerate}[topsep=0pt, 
partopsep=0pt, itemsep=0pt,parsep=0pt]
      \item 90.5(Acc.); 87.94(F score)
      \item 97.8 (Fighting activity) (F-score) \vspace{-1em}
\end{enumerate}
 & \setlist{nolistsep,leftmargin=*} \begin{enumerate}[topsep=0pt, partopsep=0pt, 
itemsep=0pt,parsep=0pt]
      \item S
      \item S  \vspace{-1em}
\end{enumerate} 
& \setlist{nolistsep,leftmargin=*} \begin{enumerate}[topsep=0pt, partopsep=0pt, 
itemsep=0pt,parsep=0pt]
      \item CS (4 subjects training, 1 subjects validation, 5 subjects test)
      \item LOSubO CV \vspace{-1em}
\end{enumerate} \\
\hline

Falling Event ~\cite{Zhang2012a} & structure-motion~\cite{Zhang2012a} & 98(insufficient illumination) \&100(sufficient illumination) & S & 
50 samples training, rest 100 test\\
\hline
UPCV ~\cite{Theodorakopoulos2014} & DS-SRC+DTW 
dissimilarity on annotated UPCV~\cite{Theodorakopoulos2014} & 89.25 & S & 
LOSubO\\
\hline
DHA ~\cite{Lin2012} & 
\setlist{nolistsep,leftmargin=*} \begin{enumerate}[topsep=0pt, partopsep=0pt, 
itemsep=0pt,parsep=0pt]
	  \item MMJRR~\cite{Gao2015}
      \item CHCRF~\cite{Liu2014a}
      \item DMPP\_PHOG~\cite{Gao2015}
      \item DLRMPP\_PHOG~\cite{Gao2015} \vspace{-1em}
\end{enumerate} & \setlist{nolistsep,leftmargin=*} \begin{enumerate}[topsep=0pt, 
partopsep=0pt, itemsep=0pt,parsep=0pt]
	  \item 98.2
      \item 95.9
      \item 95
      \item 95.6 \vspace{-1em}
\end{enumerate}
 & \setlist{nolistsep,leftmargin=*} \begin{enumerate}[topsep=0pt, partopsep=0pt, 
itemsep=0pt,parsep=0pt]
	  \item C,D
      \item C,D
      \item D
      \item C \vspace{-1em}
\end{enumerate} 
& \setlist{nolistsep,leftmargin=*} \begin{enumerate}[topsep=0pt, partopsep=0pt, 
itemsep=0pt,parsep=0pt]
	  \item LOSubO CV
      \item CS (10 training,11 test)
      \item LOSubO CV
      \item LOSubO CV \vspace{-1em}
\end{enumerate} \\
\hline
WorkoutSU-10 ~\cite{Negin2013} & 
\setlist{nolistsep,leftmargin=*} \begin{enumerate}[topsep=0pt, partopsep=0pt, 
itemsep=0pt,parsep=0pt]
	  \item Graph Mining~\cite{Celiktutan2013}
      \item Hyper-graph~\citep{Celiktutan2015} \vspace{-1em}
\end{enumerate} & \setlist{nolistsep,leftmargin=*} \begin{enumerate}[topsep=0pt, 
partopsep=0pt, itemsep=0pt,parsep=0pt]
	  \item 99.6
      \item 99.5 \vspace{-1em}
\end{enumerate}
 & \setlist{nolistsep,leftmargin=*} \begin{enumerate}[topsep=0pt, partopsep=0pt, 
itemsep=0pt,parsep=0pt]
	  \item S
      \item S \vspace{-1em}
\end{enumerate} 
& \setlist{nolistsep,leftmargin=*} \begin{enumerate}[topsep=0pt, partopsep=0pt, 
itemsep=0pt,parsep=0pt]
	  \item CS(6 subjects training, 6 test) CV
      \item CS(6 subjects training, 6 test) CV \vspace{-1em}
\end{enumerate} \\
\hline

IAS-lab ~\cite{Munaro2013, Munaro2013a}
& \setlist{nolistsep,leftmargin=*} \begin{enumerate}[topsep=0pt, partopsep=0pt, 
itemsep=0pt,parsep=0pt]
      \item SUMFLOW+PCA~\cite{Munaro2013a}
      \item Skeleton joint position~\cite{Munaro2013a} \vspace{-1em}
\end{enumerate} & \setlist{nolistsep,leftmargin=*} \begin{enumerate}[topsep=0pt, 
partopsep=0pt, itemsep=0pt,parsep=0pt]
      \item 85.2
      \item 76.7 \vspace{-1em}
\end{enumerate}
 & \setlist{nolistsep,leftmargin=*} \begin{enumerate}[topsep=0pt, partopsep=0pt, 
itemsep=0pt,parsep=0pt]
      \item C,D
      \item S \vspace{-1em}
\end{enumerate} 
& \setlist{nolistsep,leftmargin=*} \begin{enumerate}[topsep=0pt, partopsep=0pt, 
itemsep=0pt,parsep=0pt]
      \item LOSubO
      \item LOSubO \vspace{-1em}
\end{enumerate} \\
\hline
Osaka ~\cite{Mansur2013} & Dynamic features~\cite{Mansur2013} & 77.5 & S & LOSubO CV\\
\hline
Mivia ~\cite{Carletti2013} &
\setlist{nolistsep,leftmargin=*} \begin{enumerate}[topsep=0pt, partopsep=0pt, 
itemsep=0pt,parsep=0pt]      
      \item Edit distance(HARED)~\cite{Brun2015}
      \item Deep learning~\cite{Foggia2014} \vspace{-1em}  
\end{enumerate} & \setlist{nolistsep,leftmargin=*} \begin{enumerate}[topsep=0pt, 
partopsep=0pt, itemsep=0pt,parsep=0pt]
      \item 85.2 
      \item 84.7 \vspace{-1em}
\end{enumerate}
 & \setlist{nolistsep,leftmargin=*} \begin{enumerate}[topsep=0pt, partopsep=0pt, 
itemsep=0pt,parsep=0pt]
      \item D
      \item D \vspace{-1em}
\end{enumerate} 
& \setlist{nolistsep,leftmargin=*} \begin{enumerate}[topsep=0pt, partopsep=0pt, 
itemsep=0pt,parsep=0pt]
      \item LOSubO CV
      \item LOSubO CV \vspace{-1em}
\end{enumerate} \\
\hline
Concurrent Action~\cite{Wei2013a} & 
\setlist{nolistsep,leftmargin=*} \begin{enumerate}[topsep=0pt, partopsep=0pt, 
itemsep=0pt,parsep=0pt]
      \item COA~\cite{Wei2013a}    
      \item MIP~\cite{Wei2013a}
      \item Actionlet Esemble~\cite{Wang2012} \vspace{-1em}
\end{enumerate} & \setlist{nolistsep,leftmargin=*} \begin{enumerate}[topsep=0pt, 
partopsep=0pt, itemsep=0pt,parsep=0pt]
      \item 88
      \item 86
      \item 84 \vspace{-1em}
\end{enumerate}
 & \setlist{nolistsep,leftmargin=*} \begin{enumerate}[topsep=0pt, partopsep=0pt, 
itemsep=0pt,parsep=0pt]
      \item S
      \item S
      \item S \vspace{-1em}
\end{enumerate} 
& \setlist{nolistsep,leftmargin=*} \begin{enumerate}[topsep=0pt, partopsep=0pt, 
itemsep=0pt,parsep=0pt]
      \item Not given
      \item Not given
      \item Not given \vspace{-1em}
\end{enumerate} \\
\hline
3D Online ~\cite{Yu2015} & 
\setlist{nolistsep,leftmargin=*} \begin{enumerate}[topsep=0pt, partopsep=0pt, 
itemsep=0pt,parsep=0pt]
      \item Orderlets+Boosting~\cite{Yu2015}
      \item Orderlets~\cite{Yu2015}
      \item Orderlets~\cite{Yu2015} \vspace{-1em}
\end{enumerate} & \setlist{nolistsep,leftmargin=*} \begin{enumerate}[topsep=0pt, 
partopsep=0pt, itemsep=0pt,parsep=0pt]
      \item 71.4
      \item 66.1
      \item 56.4 \vspace{-1em}
\end{enumerate}
 & \setlist{nolistsep,leftmargin=*} \begin{enumerate}[topsep=0pt, partopsep=0pt, 
itemsep=0pt,parsep=0pt]
      \item S
      \item S
      \item S \vspace{-1em}
\end{enumerate} 
& \setlist{nolistsep,leftmargin=*} \begin{enumerate}[topsep=0pt, partopsep=0pt, 
itemsep=0pt,parsep=0pt]
      \item Same-Environment (2-fold CV)
      \item Cross-Environment (S1, S2 training, S3 test)
      \item Continuous (S1, S2, S3 training, S4 test) \vspace{-1em}
\end{enumerate} \\
\hline
MAD ~\cite{Huang2014} & Event transition~\cite{Kim2015} &  85.0(Frame-level Prec.); 71.41(Frame-level Rec.); 77.41(Frame-level F-score); 74.4(Event-level Prec.); 85.02(Event-level Rec.); 78.83(Event-level F-score); & S &  5-fold CV (8 groups training, 2 groups test)\\
\hline
Composable ~\cite{Lillo2014} & Hierarchical 
model~\cite{Lillo2014} & 85.7 & S & LOSubO \\
\hline
RGBD-SAR ~\cite{Zhao2013} & 
\setlist{nolistsep,leftmargin=*} \begin{enumerate}[topsep=0pt, partopsep=0pt, 
itemsep=0pt,parsep=0pt]
      \item LDP~\cite{Zhao2013}
      \item DLMC-STIPS~\cite{Ni2011} \vspace{-1em}
\end{enumerate} & \setlist{nolistsep,leftmargin=*} \begin{enumerate}[topsep=0pt, 
partopsep=0pt, itemsep=0pt,parsep=0pt]
      \item 83.5
      \item 80.2 \vspace{-1em}
\end{enumerate}
 & \setlist{nolistsep,leftmargin=*} \begin{enumerate}[topsep=0pt, partopsep=0pt, 
itemsep=0pt,parsep=0pt]
      \item C,D
      \item D \vspace{-1em}
\end{enumerate} 
& \setlist{nolistsep,leftmargin=*} \begin{enumerate}[topsep=0pt, partopsep=0pt, 
itemsep=0pt,parsep=0pt]
      \item Not given
      \item Not given \vspace{-1em}
\end{enumerate} \\
\hline
SYSU ~\cite{Hu2015} &  DS+DCP+DDP+JOULE-SVM~\cite{Hu2015} & 84.89 \& 79.63 & C,D,S & Half sample training, rest test \& CS\\
\hline 
RGB-D activity ~\cite{Wu2015} & CaTM~\cite{Wu2015} & \setlist{nolistsep,leftmargin=*} \begin{enumerate}[topsep=0pt, partopsep=0pt, 
itemsep=0pt,parsep=0pt]
      \item office: 30.6(OffSeg Acc.); 32.9(OnSeg Acc.); 33.1(OffSeg Average Prec.); 34.6(OnSeg Average Prec.); 39.9(OffFr Acc.); 38.5(OnFr Acc.); 41.5(Patching Acc.)
      \item kitchen: 33.2(OffSeg Acc.); 29.0(OnSeg Acc.); 26.4(OffSeg Average Prec.); 25.5(OnSeg Average Prec.); 37.5(OffFr Acc.); 34.0(OffFr Acc.); 20.5(Patching Acc.) \vspace{-1em}
\end{enumerate} & C,D,S & Training and test sets are specified by the author\\
\hline
UTD-MHAD ~\cite{Chen2015b} & DMM+CRC~\cite{Chen2015b} & \setlist{nolistsep,leftmargin=*} \begin{enumerate}[topsep=0pt, 
partopsep=0pt, itemsep=0pt,parsep=0pt]
      \item 79.1
      \item 67.2
      \item 66.1 \vspace{-1em}
\end{enumerate}
 & \setlist{nolistsep,leftmargin=*} \begin{enumerate}[topsep=0pt, partopsep=0pt, 
itemsep=0pt,parsep=0pt]
      \item D,I
      \item I
      \item D \vspace{-1em}
\end{enumerate}
& CS (odd indexed subjects training, rest test) \\
\hline
Morning-Routine ~\cite{Karg2013} & 
HHMM~\cite{Karg2013} & 77.01 & D
& Not given \\
\hline 
\captionsetup{width=\textwidth, belowskip=10pt,aboveskip=5pt}
\caption{Summary of state-of-the-art results with corresponding methods and settings on single-view action/activity datasets. Notation for data format: C: Colour, D: Depth, S: Skeleton, I: Inertial sensor signal.  Notation for evaluation protocol: CS: Cross subject, LOSubO: Leave one subject out, CV: Cross validation. Notation for evaluation metric: Acc.: Accuracy, Prec.: Precision, Rec.: Recall, OffSeg: Offline Segmentation, OnSeg: Online Segmentation, OffFr: Offline Frame, OnFr: Online Frame.}
\label{table:ResultsSingle}
\end{longtable}

%\begin{table*}[!htbp]
%\begin{center}
%\begin{tabular}{|p{2.5cm}|p{2cm}|p{3.5cm}|p{2.5cm}|p{1.2cm}|p{3.5cm}|}
%\scriptsize
\begin{longtable}[HTBP]{|p{1.65cm}|P{4.1cm}|P{3.8cm}|P{1.1cm}|P{4.65cm}|}
\hline
Dataset & State-of-the-art Methods 
& Acc.(\%) & Data & Protocol\\
\hline\hline
Berkeley MHAD ~\cite{Ofli2013} & 
\setlist{nolistsep,leftmargin=*} \begin{enumerate}[topsep=0pt, partopsep=0pt, 
itemsep=0pt,parsep=0pt]
	  \item Hierarchy of LDSs(28 joints used)~\cite{Chaudhry2013}
	  \item HBRNN-L(35 joints used)~\cite{Du2015}
      \item CNN(3 joints used)~\cite{Ijjina2014}
      \item Feature-Level-Fusion+SRC (Kinect+Acc1\&Acc4)~\cite{Chen2015a}
      \item HACK~\cite{Brun2014} \vspace{-1em}
\end{enumerate} & \setlist{nolistsep,leftmargin=*} \begin{enumerate}[topsep=0pt, 
partopsep=0pt, itemsep=0pt,parsep=0pt]
      \item 100
      \item 100
      \item 98.28
      \item 99.54
      \item 97.7 \vspace{-1em}
\end{enumerate}
 & \setlist{nolistsep,leftmargin=*} \begin{enumerate}[topsep=0pt, partopsep=0pt, 
itemsep=0pt,parsep=0pt]
      \item S   
      \item S
      \item S  
      \item D, A
      \item D \vspace{-1em}
\end{enumerate} 
& \setlist{nolistsep,leftmargin=*} \begin{enumerate}[topsep=0pt, partopsep=0pt, 
itemsep=0pt,parsep=0pt]
      \item CS (First 7 training, last 5 test)
      \item CS (First 7 training, last 5 test)
      \item 5-fold group-wise CV
      \item LOSubO
      \item LOSubO  \vspace{-1em}
\end{enumerate} \\
\hline
ATC$4^2$ ~\cite{Cheng2012} & 
\setlist{nolistsep,leftmargin=*} \begin{enumerate}[topsep=0pt, partopsep=0pt, 
itemsep=0pt,parsep=0pt]
      \item Depth-VSFR(All-view)~\cite{Cho2015}
      \item Depth-VSFR(Cross-view)~\cite{Cho2015}
      \item SSM(All-view)~\cite{Lee2014}
      \item SSM(Cross-view)~\cite{Lee2014} \vspace{-1em}
\end{enumerate} & \setlist{nolistsep,leftmargin=*} \begin{enumerate}[topsep=0pt, 
partopsep=0pt, itemsep=0pt,parsep=0pt]
      \item 85.5
      \item 82.0
      \item 83.4
      \item 81.2 \vspace{-1em}
\end{enumerate}
 & \setlist{nolistsep,leftmargin=*} \begin{enumerate}[topsep=0pt, partopsep=0pt, 
itemsep=0pt,parsep=0pt]
      \item D
      \item D
      \item D
      \item D \vspace{-1em}
\end{enumerate} 
& \setlist{nolistsep,leftmargin=*} \begin{enumerate}[topsep=0pt, partopsep=0pt, 
itemsep=0pt,parsep=0pt]
      \item LOSubO CV
      \item Cross-View (Training on one viewpoint, test on other viewpoints)
      \item CS (15 training,5 test,10 fold CV)
      \item CS (15 training,5 test,10 fold CV) \vspace{-1em}
\end{enumerate} \\
\hline
Falling Detection~\cite{Zhang2012b} & Bayesian framework~\cite{Zhang2012b} & 92.3(Prec.) 100(Rec.) & D & Cross-view\\
\hline
Multiview 3D Event ~\cite{Wei2013} & 
4DHOI~\cite{Wei2013} & 87
 & C, D, S
& Not given \\
\hline

Multi-View TJU ~\cite{Liu2015a} & MTSL+LL/ML~\cite{Liu2015a}& \setlist{nolistsep,leftmargin=*} \begin{enumerate}[topsep=0pt, 
partopsep=0pt, itemsep=0pt,parsep=0pt]
      \item 93.9(multi view); 91.4(front view); 90.7(side view)
      \item 95.8(multi view); 94.6(front view); 92.5(side view) \vspace{-1em}
\end{enumerate}
 & \setlist{nolistsep,leftmargin=*} \begin{enumerate}[topsep=0pt, partopsep=0pt, 
itemsep=0pt,parsep=0pt]
      \item D, S
      \item C, S \vspace{-1em}
\end{enumerate} 
& \setlist{nolistsep,leftmargin=*} \begin{enumerate}[topsep=0pt, partopsep=0pt, 
itemsep=0pt,parsep=0pt]
      \item CS(6 subjects training, 6 validation, 8 test)
      \item CS(6 subjects training, 6 validation, 8 test) \vspace{-1em}
\end{enumerate} \\
\hline

Northwestern-UCLA ~\cite{Wang2014} & \setlist{nolistsep,leftmargin=*} \begin{enumerate}[topsep=0pt, partopsep=0pt, 
itemsep=0pt,parsep=0pt]
      \item MST-AOG~\cite{Wang2014}
      \item MST-AOG~\cite{Wang2014}
      \item MST-AOG~\cite{Wang2014}
      \item NKTM~\cite{Rahmani2015}
      \item NKTM~\cite{Rahmani2015}
      \item NKTM~\cite{Rahmani2015} \vspace{-1em}
\end{enumerate} & \setlist{nolistsep,leftmargin=*} \begin{enumerate}[topsep=0pt, 
partopsep=0pt, itemsep=0pt,parsep=0pt]
      \item 81.6 
      \item 79.3
      \item 73.3
      \item 75.8
      \item 73.3
      \item 59.1 \vspace{-1em}
\end{enumerate}
 & \setlist{nolistsep,leftmargin=*} \begin{enumerate}[topsep=0pt, partopsep=0pt, 
itemsep=0pt,parsep=0pt]
      \item C, S
      \item C, S
      \item C, S
      \item C
      \item C
      \item C \vspace{-1em}
\end{enumerate} 
& \setlist{nolistsep,leftmargin=*} \begin{enumerate}[topsep=0pt, partopsep=0pt, 
itemsep=0pt,parsep=0pt]
      \item LOSubO
      \item Cross-environment
      \item Cross-view(1,2 Camera training,3 Camera test)
      \item Cross-view(1,2 Camera training,3 Camera test)
      \item Cross-view(1,3 Camera training,2 Camera test)
      \item Cross-view(2,3 Camera training,1 Camera test) \vspace{-1em}
\end{enumerate} \\
\hline

UWA3D Multiview ~\cite{Rahmani2014} & 
\setlist{nolistsep,leftmargin=*} \begin{enumerate}[topsep=0pt, partopsep=0pt, 
itemsep=0pt,parsep=0pt]
      \item Holistic HOPC(Same view)~\cite{Rahmani2014}
      \item MSO-SVM(Cross view)~\cite{Rahmani2014} \vspace{-1em}
\end{enumerate} & \setlist{nolistsep,leftmargin=*} \begin{enumerate}[topsep=0pt, 
partopsep=0pt, itemsep=0pt,parsep=0pt]
      \item 84.93 
      \item 91.79($0^{\circ}$), 86.67($-25^{\circ}$), 88.89($+25^{\circ}$), 
75.56($-50^{\circ}$), 77.78($+50^{\circ}$) \vspace{-1em}
\end{enumerate}
 & \setlist{nolistsep,leftmargin=*} \begin{enumerate}[topsep=0pt, partopsep=0pt, 
itemsep=0pt,parsep=0pt]
      \item D
      \item D \vspace{-1em}
\end{enumerate} 
& \setlist{nolistsep,leftmargin=*} \begin{enumerate}[topsep=0pt, partopsep=0pt, 
itemsep=0pt,parsep=0pt]
      \item CS (Half training, half test) 
      \item $0^{\circ}$ training \vspace{-1em}
\end{enumerate} \\
\hline
NJUST~\cite{Song2014} & 
\setlist{nolistsep,leftmargin=*} \begin{enumerate}[topsep=0pt, partopsep=0pt, 
itemsep=0pt,parsep=0pt]
      \item ToSP+SVM~\cite{Song2014}
      \item BSC+Spatial-Temporal~\cite{Song2015} \vspace{-1em}
\end{enumerate} & \setlist{nolistsep,leftmargin=*} \begin{enumerate}[topsep=0pt, 
partopsep=0pt, itemsep=0pt,parsep=0pt]
      \item 98.4
      \item 94.7 \vspace{-1em}
\end{enumerate}
 & \setlist{nolistsep,leftmargin=*} \begin{enumerate}[topsep=0pt, partopsep=0pt, 
itemsep=0pt,parsep=0pt]
      \item C,D     
      \item D \vspace{-1em}
\end{enumerate} 
& \setlist{nolistsep,leftmargin=*} \begin{enumerate}[topsep=0pt, partopsep=0pt, 
itemsep=0pt,parsep=0pt]
      \item LOSubO
      \item LOSubO \vspace{-1em}
\end{enumerate} \\
\hline
DMLSmart Actions ~\cite{Amiri2013} & 
\setlist{nolistsep,leftmargin=*} \begin{enumerate}[topsep=0pt, partopsep=0pt, 
itemsep=0pt,parsep=0pt]
      \item SVM-NNSC + Proposed Kernel~\cite{Amiri2014a}
      \item Meta Learning~\cite{Amiri2014} \vspace{-1em}
\end{enumerate} & \setlist{nolistsep,leftmargin=*} \begin{enumerate}[topsep=0pt, 
partopsep=0pt, itemsep=0pt,parsep=0pt]
      \item 79.9
      \item 77.19 \vspace{-1em}
\end{enumerate}
 & \setlist{nolistsep,leftmargin=*} \begin{enumerate}[topsep=0pt, partopsep=0pt, 
itemsep=0pt,parsep=0pt]
      \item HDC     
      \item C,D \vspace{-1em}
\end{enumerate} 
& \setlist{nolistsep,leftmargin=*} \begin{enumerate}[topsep=0pt, partopsep=0pt, 
itemsep=0pt,parsep=0pt]
      \item LOSubO
      \item LOSubO \vspace{-1em}
\end{enumerate} \\
\hline
ReadingAct ~\cite{Chen2013} & 
\setlist{nolistsep,leftmargin=*} \begin{enumerate}[topsep=0pt, partopsep=0pt, 
itemsep=0pt,parsep=0pt]
      \item BoW+$\chi^2$ SVM~\cite{Chen2013}
      \item BoW+Linear SVM~\cite{Chen2013} \vspace{-1em}
\end{enumerate} & \setlist{nolistsep,leftmargin=*} \begin{enumerate}[topsep=0pt, 
partopsep=0pt, itemsep=0pt,parsep=0pt]
      \item 90.4
      \item 82.1 \vspace{-1em}
\end{enumerate}
 & \setlist{nolistsep,leftmargin=*} \begin{enumerate}[topsep=0pt, partopsep=0pt, 
itemsep=0pt,parsep=0pt]
      \item C
      \item C,D \vspace{-1em}
\end{enumerate} 
& \setlist{nolistsep,leftmargin=*} \begin{enumerate}[topsep=0pt, partopsep=0pt, 
itemsep=0pt,parsep=0pt]
      \item CS (15 training,5 test,4-fold CV)
      \item CS (15 training,5 test,4-fold CV) \vspace{-1em}
\end{enumerate} \\
\hline
\captionsetup{width=\textwidth}
\caption{Summary of state-of-the-art results with corresponding methods and protocols on multi-view action /activity datasets. 
Notation for data format: C: Colour, D: Depth, S: Skeleton, A: Acceleration, HDC: High Definition Colour. Notation for evaluation protocol: CS: Cross subject, LOSubO: Leave one subject out, CV: Cross validation. Notation for evaluation metric: Acc.: Accuracy.}
\label{table:ResultsMultiview}
\end{longtable}
%\end{tabular}
%\end{center}
%\caption{Summary of the key specifications of the multi-view action and activity 
%datasets and state-of-the-art results with corresponding methods and protocols. 
%The acronyms are same as the ones in Table~\ref{table:Basicsingle}}
%\label{table:Basicmultiview}
%\end{table*}

%\begin{table*}[!htbp]
%\begin{center}
%\begin{tabular}{|p{2.5cm}|p{2cm}|p{3.5cm}|p{2.5cm}|p{1.2cm}|p{3.5cm}|}
%\scriptsize
\begin{longtable}[HTBP]{|p{2.5cm}|P{3.8cm}|P{3.8cm}|P{1.3cm}|P{3.8cm}|}
\hline
Dataset & State-of-the-art Methods 
& Acc.(\%) & Data & Protocol\\
\hline\hline
SBU ~\cite{Yun2012}  & 
\setlist{nolistsep,leftmargin=*} \begin{enumerate}[topsep=0pt, partopsep=0pt, 
itemsep=0pt,parsep=0pt]
      \item MaxEnt IOC~\cite{Huang2015}
      \item DMDP~\cite{Huang2015, Kitani2012} \vspace{-1em}
\end{enumerate} & \setlist{nolistsep,leftmargin=*} \begin{enumerate}[topsep=0pt, 
partopsep=0pt, itemsep=0pt,parsep=0pt]
      \item 0.52 (AFD); 80 (NLL)
      \item 0.51 (AFD); 113.5 (NLL) \vspace{-1em}
\end{enumerate}
 & \setlist{nolistsep,leftmargin=*} \begin{enumerate}[topsep=0pt, partopsep=0pt, 
itemsep=0pt,parsep=0pt]
      \item S
      \item S \vspace{-1em}
\end{enumerate} 
& \setlist{nolistsep,leftmargin=*} \begin{enumerate}[topsep=0pt, partopsep=0pt, 
itemsep=0pt,parsep=0pt]
      \item LOSubO (7-fold CV)
      \item LOSubO (7-fold CV) \vspace{-1em}
\end{enumerate} \\
\hline
K3HI ~\cite{Hu2013} & Positive action (Joint 
motion)~\cite{Hu2013} & 75.6 & S
& 4-fold CV \\
\hline
LIRIS ~\cite{Wolf2014} & 
\setlist{nolistsep,leftmargin=*} \begin{enumerate}[topsep=0pt, partopsep=0pt, 
itemsep=0pt,parsep=0pt]
      \item Pose+Appearance +Context+Scene (With Localization)~\cite{Ni2013, 
Wolf2014}
      \item Pose+Appearance +Context+Scene (Without Localization)~\cite{Ni2013, 
Wolf2014} \vspace{-1em}
\end{enumerate} & \setlist{nolistsep,leftmargin=*} \begin{enumerate}[topsep=0pt, 
partopsep=0pt, itemsep=0pt,parsep=0pt]
      \item 74(Rec.); 41(Prec.); 53(F-score)
      \item 63(Rec.); 33(Prec.); 44(F-score) \vspace{-1em}
\end{enumerate}
 & \setlist{nolistsep,leftmargin=*} \begin{enumerate}[topsep=0pt, partopsep=0pt, 
itemsep=0pt,parsep=0pt]
      \item G, D
      \item G, D \vspace{-1em}
\end{enumerate} 
& \setlist{nolistsep,leftmargin=*} \begin{enumerate}[topsep=0pt, partopsep=0pt, 
itemsep=0pt,parsep=0pt]
      \item 305 action samples training, 156 samples test
      \item 305 action samples training, 156 samples test \vspace{-1em}
\end{enumerate} \\
\hline
G3Di ~\cite{Bloom2014} & 
\setlist{nolistsep,leftmargin=*} \begin{enumerate}[topsep=0pt, partopsep=0pt, 
itemsep=0pt,parsep=0pt]
      \item Action segment~\cite{Bloom2014}
      \item Action points~\cite{Bloom2013}
\end{enumerate} & \setlist{nolistsep,leftmargin=*} \begin{enumerate}[topsep=0pt, 
partopsep=0pt, itemsep=0pt,parsep=0pt]
      \item  Action: 56.1(F1); Interaction: 57.1(F1) 
      \item  Action: 42.6(F1); Interaction: 44.8(F1) \vspace{-1em}
\end{enumerate}
 & \setlist{nolistsep,leftmargin=*} \begin{enumerate}[topsep=0pt, partopsep=0pt, 
itemsep=0pt,parsep=0pt]
      \item S (Boxing)
      \item S (Boxing) \vspace{-1em}
\end{enumerate} 
& \setlist{nolistsep,leftmargin=*} \begin{enumerate}[topsep=0pt, partopsep=0pt, 
itemsep=0pt,parsep=0pt]
      \item LOSubO
      \item LOSubO \vspace{-1em}
\end{enumerate} \\
\hline
Office Activity~\cite{Wang2014a} & Structured deep architecture~\cite{Wang2014a} & 60.1(OA1); 45.0(OA2) & D & 5-fold CV\\
\hline
$M^2$I TJU ~\cite{Xu2015} & - & - & - & -\\
\hline
ShakeFive~\cite{Gemeren2014} & 
\setlist{nolistsep,leftmargin=*} \begin{enumerate}[topsep=0pt, partopsep=0pt, 
itemsep=0pt,parsep=0pt]
      \item Dyadic poselets~\cite{Gemeren2014}
      \item Dyadic poselets~\cite{Gemeren2014} \vspace{-1em}
\end{enumerate} & \setlist{nolistsep,leftmargin=*} \begin{enumerate}[topsep=0pt, 
partopsep=0pt, itemsep=0pt,parsep=0pt]
      \item 49.56 (Handshake) 34.85 (Highfive)
      \item 47.87 (Handshake) 23.94 (Highfive) \vspace{-1em}
\end{enumerate}
 & \setlist{nolistsep,leftmargin=*} \begin{enumerate}[topsep=0pt, partopsep=0pt, 
itemsep=0pt,parsep=0pt]
      \item C, S
      \item C, S \vspace{-1em}
\end{enumerate}
& \setlist{nolistsep,leftmargin=*} \begin{enumerate}[topsep=0pt, partopsep=0pt, 
itemsep=0pt,parsep=0pt]
      \item 75\% training (4-fold CV)
      \item 25\% training (4-fold CV) \vspace{-1em}
\end{enumerate} \\
\hline
\captionsetup{width=\textwidth}
\caption{Summary of state-of-the-art results with corresponding methods and protocols on human-human interaction and multi-person action/activity datasets. Notation for data format: C: Colour, D: Depth, S: Skeleton, G: Grayscale. Notation for evaluation protocol: LOSubO: Leave one subject out, CV: Cross validation. Notation for evaluation metric: Acc.: Accuracy, Prec.: Precision, Rec.: Recall, AFD: Average image Feature Distance, NLL: Negative Log-Likelihood.}
\label{table:ResultsMultiperson}
\end{longtable}
\end{scriptsize}

\subsection{Recommendations}
\label{sec:Recommendations}
The intensity of research activity in human action/activity 
recognition has encouraged the development of new algorithms and 
possibly the generation of new datasets. Based on our review, some 
newly collected datasets share similar characteristics with existing 
ones and may not have expanded the variety of environmental factors inherent 
in the dataset. Perhaps more importantly, comparisons between algorithms 
evaluated on different datasets are in many cases unfair and 
 makes the progress achieved to date unclear. Here, we make some 
recommendations on the issues of dataset selection and 
evaluation protocols.

%Therefore, we make some recommendations on the selection of datasets according to specific application and aspect of action/activity recognition to avoid these issues to some extent.

%In this section, some of the issues associated with the existing datasets and evaluation protocols are highlighted along with our recommendations.

%However, as we addressed, some newly collected datasets have similar properties with existing ones, which means the time and resources that are spent on collecting new datasets are wasted in this case. More importantly, comparisons between algorithms that tested on different datasets are not fair and justifiable. Therefore, we make some recommendations on the selection of datasets according to specific application and aspect of action/activity recognition to avoid these issues to some extent.
%In this section, some of the issues associated with the existing datasets and evaluation protocols are highlighted along with our recommendations.

\subsubsection{Datasets}

It is clear that each of the datasets are matched to a specific application 
and aspect of action/activity recognition. 
%Even in this situation, none was able to accommodate all the various factors already identified for the design of robust algorithms. 
Inherent in each dataset are factors that the algorithm under evaluation is 
meant to accommodate. These factors include variation of 
execution rate and style of performance, degree of clutter in background and 
occlusion, 
multi view points, camera motion, action detection, and online learning. All of 
these factors have been analysed in Section~\ref{sec:complexity}.
%, such as variation of execution rate and anthropometric, and cluttered background and occlusion, which allows us to provide the recommendations directly, and camera motion.

%Here we analyse the rest two factors related to algorithms. For the action detection, the dataset should be captured continuously and provide accurate ground truth segmentation points of each action. To evaluate the online learning algorithms, a dataset should be captured continuously and actions were performed in random order, which is slightly different from datasets testing action detection.

Based on the analysis, below, we provide the list of environmetal factors and
applications, along with the datasets that incorporate/are suitable for them as 
a guide in their selection. 
 
\begin{description} [topsep=0pt, partopsep=0pt, itemsep=0pt,parsep=0pt]
%\vspace{-0.3cm}
 %     \item{{\bf Variation of execution rate and anthropometric}} RGBD-HuDaAct (30 subjects involved), RGBD-SAR (30 subjects involved), MSRC-12(30 subjects involved), ATC$4^2$ (24 subjects involved), Multiview 3D Event (Around 20 repetitions per subject), and LIRIS (21 subjects involved). \vspace{-0.3cm}
 		\item [Execution rate and anthropomorphic variation:] 
RGBD-HuDaAct, MSRC-12, Concurrent action, RGBD-SAR, composable, DMLSmart, 
Multiview 3D Event, LIRIS, and Office Activity. 
%\vspace{-0.3cm}
      \item [Cluttered background and occlusion:] UTKinect, RGBD-HuDaAct, 
MSRDaily Activity, CAD-120, RGBD-SAR, 3D Online, DMLSmartActions, Multiview 3D 
Event, LIRIS, and Office Activity. 
%\vspace{-0.3cm}
      \item [Multi viewpoints:] ATC$4^2$, Falling Detection, Berkeley 
MHAD, DMLSmartActions, ReadingAct, Multiview 3D Event, Northwestern-UCLA 
Multiview, UWA3D Multiview, Multi-view TJU, NJUST, $M^2$I TJU, and Office Activity. 
%\vspace{-0.3cm}
      \item [Camera motion:] LIRIS. 
      %\vspace{-0.3cm}
%      \item Object tracking: \\
%      CAD-120, 3D Online and Multiview3DEvent. \vspace{-0.3cm}
      \item [Action detection:] CAD-60, CAD-120, MAD, Human Morning Routine, 
Composable, Multiview 3D Event, and G3Di.
%\vspace{-0.3cm}
      \item [Falling detection:] Falling Detection, Falling Event, and 
ATC$4^2$.
      \item [Online action recognition:] 3D Online, Concurrent Action,  RGB-D activity,
DMLSmartActions.
      \item [Object detection:] CAD-120, Human Morning Routine, 3D Online, and 
Multiview 3D Event.
%\vspace{-0.3cm}
      
\end{description}

\subsubsection{Evaluation protocols}
This review suggests that the most widely adopted experimental set up in 
the state-of-the-art results are ``leave-one-subject-out cross validation'' 
and ``cross-subject test''.  The fact that several datasets are 
released without an accompanying \textit{de facto} standard evaluation protocol 
results in controversial comparisons among algorithms. For example, the 
summaries of evaluation protocols given in 
section~\ref{sec:evaluationProtocol} shows that the most commonly used 
cross-subject scheme has different splitting methods. Some papers used odd 
indexed subjects as training and even indexed subjects as test, others may use 
first half of subjects as training data and the remainder as test data. Some 
have used cross-validation on the split data and some have only reported the 
results on one test. There are some papers that did not provide 
explicit information on the evaluation protocol used.

We recommend that any new release of dataset should be accompanied by 
``standard'' and unified evaluation protocols, that future proposed 
algorithms can use for design and performance evaluation. Admittedly, some 
applications may require specific evaluation methods different from those 
published with a given dataset. New evaluation protocols should be clearly 
articulated and provided with informative justification.
%The most widely adopted experimental settings are leave-one-
% subject-out cross validation set-up and cross-subject set-up. However, these 
% settings are not without controversy. In most of the datasets, the cameras are 
% fixed and background would not have changed during data capture. Furthermore, 
% within a specific dataset the rules for performing the actions are fixed and all 
% subjects usually performed actions from a fixed location in a scene. These 
% issues may limit the robustness of algorithms if cross-subject or 
% leave-one-subject-out cross validation schemes are used. One reason adduced for 
% this limitation is that algorithms may inadvertently rely on the background 
% information or the position of actors. In this case, if new datasets have 
% different distribution with training data need to be tested, even though the new 
% dataset shares some similar actions as the existing one, all the actions need to 
% be labelled again and the model needs to be retrained. From this perspective, 
% these algorithms may not be robust enough for real world applications. 

% %To overcome the drawbacks of cross-subject or leave-one-subject-out 
% cross validation schemes, we recommend to use cross-dataset evaluation scheme. 
% In cross-dataset set-up, the actors, environment and manner performing actions 
% in training and test data are all different. Such a protocol requires the 
% algorithm to be robust in order to consistently achieve high recognition 
% accuracy.

\section{Discussion}
\label{sec:Discussion}
In this section, we point out the limitations of both current RGB-D 
action datasets and commonly used evaluation protocols on action recognition. 
Our aim is to provide guidance on future creation of datasets and 
establishment of standard evaluation protocols for specific purposes.

\subsection{Limitations of current datasets}
\label{sec:LimDatasets}
The review and analysis of current RGB-D action datasets have revealed 
some limitations including size, applicability, availability of ground truth 
labels and evaluation protocols. There is also the problem of dataset 
saturation, a phenomenon whereby algorithms reported have achieved a 
near-perfect performance. We now elaborate on these limitations.
% We also find some of the 
% datasets are almost saturated based on the results obtained on them. Hence, we 
% reveal and discuss the limitations of current datasets in this section.
\begin{description}[topsep=0pt, partopsep=0pt, itemsep=0pt,parsep=0pt]
\item{\textbf{Dataset size: }}The most obvious limitation of 
current dataset is the small number of action classes and sample size. 
Current RGB-D based action datasets typically contain 10 to 20 actions, which is 
not comparable to those of 2D video action datasets. A newly released 2D 
dataset~\cite{Heilbron2015} on action recognition contains 203 distinct 
action classes in 849 hours of video recording. 
% This dataset focuses on benchmarking daily life related activities. 
Another 2D dataset~\cite{Karpathy2014} on sport activities contains 1 million 
YouTube videos aggregating 487 classes. A possible reason is that it is 
relatively easy to ``harvest'' 2D videos from the Internet, e.g., YouTube. In 
contrast, the RGB-D based videos have to be captured manually and, the time, 
financial and labour constraints limit the size of RGB-D datasets. 
\item{\textbf{Applicability: }}The applications of current 
RGB-D-based action datasets are also very limited because of the restricted 
types of actions represented in each dataset. Most RGB-D datasets are 
collected within 
lab environment and the execution style of actions generally follow strict 
instructions. Thus, even with different subjects, the variations 
in performing style are subtle and indiscernible. 
\item{\textbf{Ground truth: }}Some of current datasets are 
well constructed with many challenging factors, however, they provide poor 
ground truth labels, which limits their usability.
\item{\textbf{Evaluation protocols: }}As analysed in 
Sections~\ref{sec:evaluationProtocol} and~\ref{sec:Recommendations}, the 
controversy of evaluation protocols may lead to unfair comparison among 
algorithms; a situation largely due to lack of clarity on the protocols to be 
used with published datasets.
\item{\textbf{Saturation: }}Section~\ref{sec:Recommendations} 
has provided recommendations on the selection of dataset for different 
purposes, suggesting that current datasets already represent the 
environmental factors required to rigorously test and evaluate different 
algorithms. However, based on the state-of-the-art results 
summarised in Section~\ref{sec:State-of-the-art}, it can be seen that 
algortihms have already achieved a near-perfect 
accuracy on some modalities of these datasets. This suggests that these datasets 
are near saturated. This phenomenon obscures the fact that algorithms may not 
yet be suitable for deployment in real-world applications. It is necessary 
that the set of environmental factors  and their 
level of complexities (Section~\ref{sec:complexity}), are matched to real-world 
applications and, guide the creation of new and challenging RGB-D action 
dataset.
\end{description}

\subsection{Recommendations for future datasets}
\label{sec:RecDatasets}
Based on the limitations identified above we provide some 
recommendations on creating future datasets. The number of samples and variety 
of action types needs to be increased so that a learning algorithm may 
generalize on the problem domain. Algorithms are destined for inclusion in some 
real-world applications and as such dataset creators may need to focus on 
specific applications and the inherent environmental factors. This will allow 
the creation of datasets with realistic and free-form performance of actions 
that properly model the problem. 
The proliferation of datasets has its advantage namely, opportunity to expand 
the test and evaluation suite. However, there is opportunity to create 
sequentially captured and randomly performed RGB-D action recognition dataset. 
The ground truth will then be the action segment points. Such dataset will be 
an all-in-one testing suite for different algorithms - action categorization, 
action detection and online recognition. Apart from the provision of action 
segement as ground truth, actor and object locations along with any other 
informative metadata should be provided along with the dataset.  

Finally, a dataset should be published with a number of standard 
evaluation protocols for use in the design, testing and fair 
comparative evaluation of future algorithms. Perhaps more importantly, the 
evaluation protocols should match real-world applications expectation. For 
example, in video surveillance applications, the cross-subject scheme is more 
appropriate than leave-one-sequence-out scheme. However, in health monitoring 
applications, as the system only monitors specific subject without new subjects, 
the leave-one-sequence-out scheme is more appropriate.

\subsection{Limitations of evaluation protocols}
\label{sec:LimProtocols}
Incidentally, the limitations of evaluation protocols may impede the progress 
of action recognition algorithms towards maturity and robustness for 
real-world applications. Currently, the most 
widely adopted experimental settings are leave-one-subject-out cross validation 
set-up and cross-subject set-up. However, these settings are not without 
controversy from the real-world application perspectives. In most of the 
datasets, the cameras are fixed and background would not have changed during 
data capture. Furthermore, within a specific dataset the instructions for 
performing the actions are fixed and all subjects usually performed actions from 
a fixed location in a scene. These issues may limit the robustness of algorithms 
if cross-subject or leave-one-subject-out cross validation schemes are used. 
One reason adduced for this limitation is that algorithms may inadvertently rely 
on the background information or the position of actors. Hence, the algorithms 
tested using these protocols can only be used on particular real world 
applications where the background and camera are fixed. 

% More importantly, if new datasets need to be tested, even though 
% the new dataset shares some similar actions with the existing one, all the 
% actions need to be labelled again and the model needs to be retrained to employ 
% the leave-one-subject-out or cross-subject protocols. From this perspective, 
% these protocols may not be applicable enough for real world applications. 

To some degree, the cross-view and cross-environment protocols are more 
realistic than leave-one-subject-out and cross-subject versions. These 
protocols 
consider the variation of viewpoints and surrounding environments of the 
performed actions. However, those protocols can only be used with specific 
datasets having multi-view points or multiple capture environment. Moreover, 
these protocols retain the problem associated with similar performance styles 
between training and test set. They are limited to one dataset in which 
the actions are performed under identical instructions.

\subsection{Recommendations for future evaluation protocols}
\label{sec:RecProtocols}
As mentioned in Section~\ref{sec:RecDatasets}, evaluation 
protocols should correspond to specific real-world applications. The 
cross-subject, leave-one-subject-out cross validation, cross-view, and 
cross-environment schemes can either only be used with specific datasets or 
for particular applications.

To overcome the drawbacks of current evaluation schemes, we advocate
the use of cross-dataset evaluation scheme. In a cross-dataset 
set-up, the actors, view point, environment, and manners of performing actions 
in training and test data are all different. Furthermore it is not 
limited to a specific dataset, since any group of datasets that share 
similar actions and semantics can be used. Perhaps more importantly, the 
cross-dataset evaluation scheme is more akin to real-world applications where 
the system trained on particular scenario can be used in other similar 
scenarios without the need to retrain the whole system.

The cross-dataset scheme has already been  adopted on some algorithms for 
action recognition in 2D videos~\cite{Cao2010}~\cite{Sultani2014}, however, to 
our best knowledge, there is no report of its usage on RGB-D video 
datasets. Such a protocol requires the algorithm to be robust and able to 
accommodate the various environmental factors in order to consistently perform 
well.

It is interesting to note that in the evaluation scheme it is common to report 
the average of several runs. While this is a good statistical practice, we 
notice that such avearges are compared straightforwardly with results from 
existing algorithms without a test of the statistical significance of the 
observed difference. Perhaps, in line with protocols of well designed 
statistical experiments, the results reported for action recogniton algorithms 
should also include statistical significance tests\cite{Japkowicz2014}.
% achieve high recognition accuracy. Hence, the cross-dataset 
% scheme is % challenging and open in the field of RGB-D-based action recognition, 
% which is % also a good 
% research direction in the future.

\section{Conclusion}
\label{sec:conclusion}
A comprehensive review of commonly used and publicly available 
RGB-D-based datasets for action recognition has been provided. The detailed 
descriptions and analysis, highlights of their characteristics and potential 
applications should be useful for researchers designing action recognition 
algorithms. This is especially so, when selecting datasets for 
algorithm development and evaluation as well as creating new datasets to 
fill identified gaps. Most of the datasets collected to date are meant for  
algorithms devised to solve specific action recognition problem. However, the 
simplicity of the datasets have resulted in a ``saturated'' state whereby 
algorithmic improvement has stalled. A more realistic collection of datasets 
representing a broad selection of challenging enviromental factors is now 
required.  We have advocated the use of cross-dataset evaluation set 
up to provide a more realistic testing scenario. Furthermore, we advoacted the 
use of evaluation protocol that include statistical significance test to ensure 
fair comparision amongst 
algorithms. Meanwhile, the state-of-the-art results over the datasets 
we reviewed have been provided in one place to help researchers when 
configuring their comparative evaluation schedule. We also summarise several 
commonly used evaluation and validation set-ups and address their drawbacks, 
resulting in a set of recommendations on future collection of datasets and use 
of evaluation protocols.

This review has highlighted the need for comprehensive statistically 
significant evaluation protocols as part of algorithm development and testing.  
We are working on publshing an open-source software suite that will enable 
easy evaluation of action recognition algorithms, especially with cross dataset 
schemes.

\section*{References}
\small
\bibliographystyle{elsarticle-num}
\bibliography{SurveyOfDatasets}

\end{document}